\documentclass[review]{elsarticle}

\usepackage{lineno}
\usepackage{fixltx2e}
\usepackage{times}
\usepackage{epsfig}
\usepackage{graphicx}
\usepackage{amsmath}
\usepackage{amssymb}
\usepackage{color}
\usepackage{soul}
\usepackage{placeins}
\usepackage{paralist}
\usepackage{tabularx}
\usepackage{multirow}
\usepackage{multicol}
\usepackage{subfigure}
\usepackage{enumitem}
\usepackage[usenames,dvipsnames]{xcolor}
\usepackage[pagebackref=true,breaklinks=true,letterpaper=true,colorlinks,bookmarks=false]{hyperref}
\modulolinenumbers[5]

\journal{Computer Vision and Image Understanding}

\begin{document}

\begin{frontmatter}

\title{The THUMOS Challenge on Action Recognition for \\Videos ``in the Wild''\tnoteref{mytitlenote}}
\tnotetext[mytitlenote]{www.thumos.info}


\author[ucfaddress]{Haroon Idrees\corref{mycorrespondingauthor}}
\cortext[mycorrespondingauthor]{Corresponding author}
\ead{haroon@cs.ucf.edu}

\author[stanfordaddress]{Amir R. Zamir}

\author[fudanaddress]{Yu-Gang Jiang}

\author[googleaddress]{Alex Gorban}


\author[inriaaddress]{Ivan Laptev}

\author[googleaddress]{Rahul Sukthankar}

\author[ucfaddress]{Mubarak Shah}

\address[ucfaddress]{Center for Research in Computer Vision, University of Central Florida, Orlando, USA}
\address[stanfordaddress]{Dept.\ of Computer Science, Stanford University, USA}
\address[fudanaddress]{School of Computer Science, Fudan University, Shanghai, China}
\address[inriaaddress]{INRIA Paris - Rocquencourt, France}
\address[googleaddress]{Google Research, USA}

%
%
%

\begin{abstract}
Automatically recognizing and localizing wide ranges of human actions has crucial importance for video understanding. Towards this goal, the THUMOS challenge was introduced in 2013 to serve as a benchmark for action recognition. Until then, video action recognition, including THUMOS challenge, had focused primarily on the classification of pre-segmented (i.e., trimmed) videos, which is an artificial task. In THUMOS 2014, we elevated action recognition to a more practical level by introducing temporally untrimmed videos. These also include `background videos' which share similar scenes and backgrounds as action videos, but are devoid of the specific actions. The three editions of the challenge organized in 2013--2015 have made THUMOS a common benchmark for action classification and detection and the annual challenge is widely attended by teams from around the world.

In this paper we describe the THUMOS benchmark in detail and give an overview of data collection and annotation procedures. We present the evaluation protocols used to quantify results in the two THUMOS tasks of action classification and temporal detection. We also present results of submissions to the THUMOS 2015 challenge and review the participating approaches. Additionally, we include a comprehensive empirical study evaluating the differences in action recognition between trimmed and untrimmed videos, and how well methods trained on trimmed videos generalize to untrimmed videos. We conclude by proposing several directions and improvements for future THUMOS challenges.
\end{abstract}

\begin{keyword}
Action Recognition, Action Detection, Action Localization, Untrimmed Videos, THUMOS, Dataset, Benchmark, UCF101
\end{keyword}

\end{frontmatter}


\section{Introduction}\label{sec:intro}

The action recognition community has made great progress in the last few years, driven in large part by the release of large video datasets such as UCF101~\cite{Soomro12} and HMDB~\cite{Kuehne11} in conjunction with the development of new features~\cite{Wang13}, representations~\cite{Oneata13} and learning methods~\cite{Simonyan14}. Recent datasets contain challenging videos with actions from various sources such as movies~\cite{Kuehne11,Marszalek09}, YouTube~\cite{Liu09}, and wearable cameras~\cite{Pirsiavash12,Ryoo13}. The performance of methods evaluated on such datasets has steadily increased over the years~\cite{Wang13}. In line with these advances in action recognition, the THUMOS challenge was introduced to the computer vision community in 2013 with the aim to explore and evaluate new approaches for large-scale action analysis from Internet videos in a realistic setting.

The THUMOS 2013 challenge was based on the UCF101 dataset~\cite{Soomro12}, which similar to most of the commonly evaluated action recognition datasets consists exclusively of manually trimmed video clips that exclude temporal clutter. The assumption of such clean and trimmed videos may be reasonable during training time since it provides methods with strongly supervised data. However, the same restriction during testing is potentially impractical and unreasonable for several reasons:
\begin{itemize}
\item it assumes an (unrealistic) external process to temporally segment videos into clips that precisely surround the desired action;
\item it creates a test set distribution that does not match the real-world distribution since the test data is free from temporal clutter, `background' class data notwithstanding;
\item it can allow methods to inadvertently exploit side-information, such as the length of the test video clip~\cite{Satkin10}, even though this information is available only due to an artifact of the evaluation methodology.
\end{itemize}

Thus, the temporally segmented clips do not reflect the real world as the actions are typically embedded in complex dynamic scenes with rich causal and spatial relations among people and objects. While elimination of temporal clutter simplifies the recognition problem, it becomes difficult to predict the performance of different methods in real applications. In literature, there have been some efforts to address the problem of action recognition in untrimmed videos. For example, temporal detection has been studied in~\cite{Bojanowski14,Duchenne09,Hoai11,Pirsiavash14}, while spatiotemporal localization of actions has been addressed in~\cite{Ke07,Klaser10,Laptev07,Tian13}. Such works deal with substantial amount of temporal clutter from movies and sports videos. However, they typically were evaluated on only a small number of action classes and required strongly supervised training and test sets. The THUMOS'14 challenge~\cite{THUMOS14} introduced thousands of untrimmed videos in validation, background and test sets for 101 action classes providing the community with the first-of-its-kind dataset for action recognition and temporal detection in realistic settings with a standardized evaluation protocol. Similarly, THUMOS'15 challenge~\cite{THUMOS15} extended the THUMOS'14 dataset by including a new test set constituting 5,613 positive and background untrimmed videos.

THUMOS (Greek: $\theta\upsilon\mu\acute{o}\zeta$) which means a \textit{spirited contest}, consists of two principal challenges: \textit{classification} - where the goal is to determine whether a video contains a particular action or not, and \textit{temporal detection} - where the goal is to classify an action find its temporal locations in each video. The THUMOS action classes are from UCF101 \cite{Soomro12} and can be divided into five main categories: \textit{Human-Object Interaction}, \textit{Body-Motion Only}, \textit{Human-Human Interaction}, \textit{Playing Musical Instruments}, and \textit{Sports}. All the videos are publicly available from YouTube\footnote{http://www.youtube.com/}, and manually annotated both for action label and temporal span of each action.

\begin{figure}[t]
\begin{center}
\includegraphics[width=0.70\linewidth]{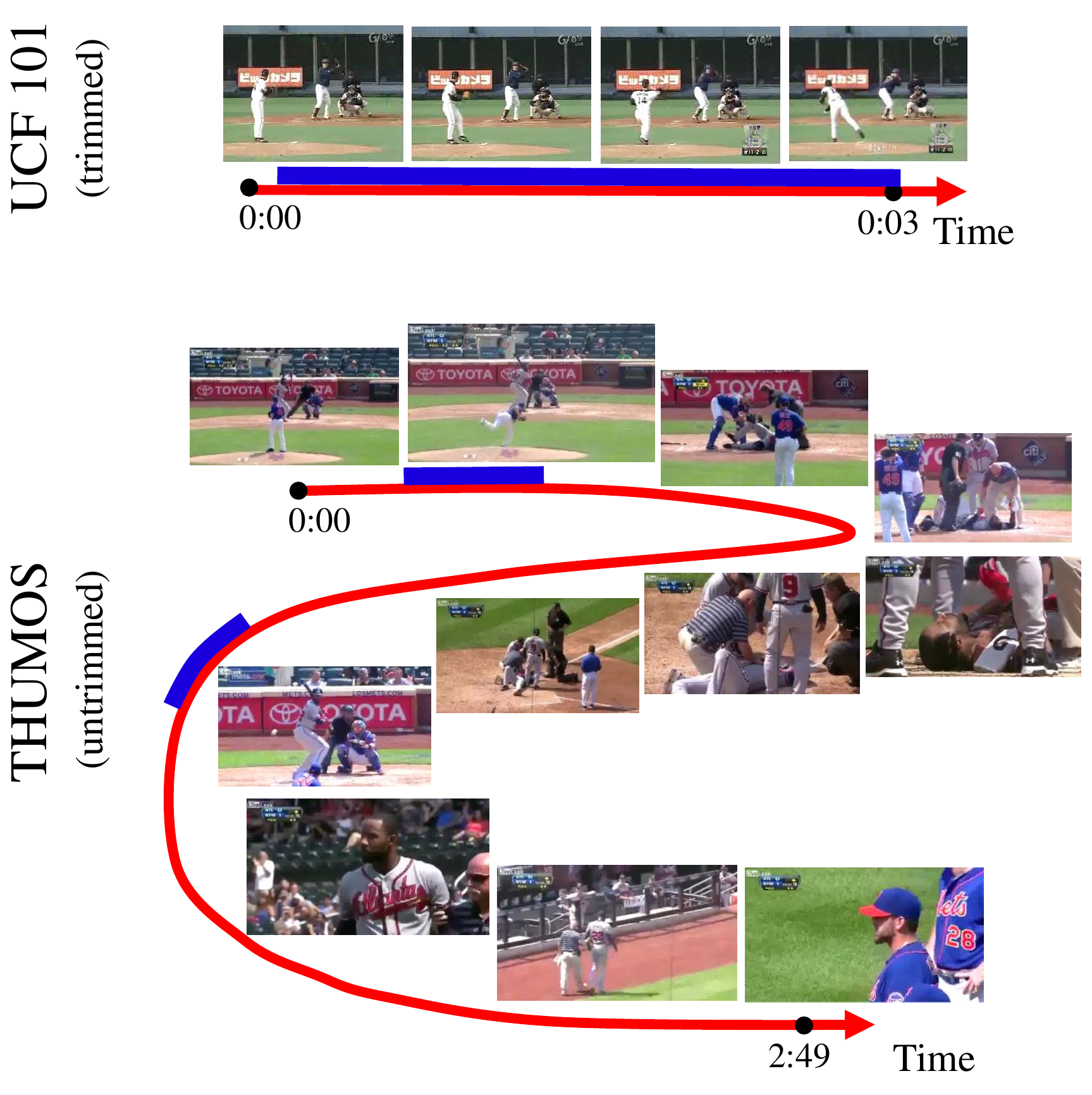}\vspace{-.3cm}\\
\end{center}
\caption{Illustration contrasting a (trimmed) video clip for the `BaseballPitch' action from the UCF101 dataset and an \textit{untrimmed video} from the corresponding action taken from the validation set of THUMOS'15. Note that the entire temporal span of the video (shown in red) contains a variety of baseball actions with the pitch occurring multiple times (shown in blue).}
\label{figTeaser}
\end{figure}

The objectives of the THUMOS challenge are twofold: a) to serve as a benchmark and enable a comparison of different approaches on the tasks of action classification and temporal detection in large-scale realistic video settings; and b) to advance the state of the art. For instance, the accuracy on UCF101 increased from 45\% in 2012 to almost 90\% at THUMOS'13~\cite{THUMOS13}. Similarly, the 2014 and 2015 challenges are characterized by three significant differences compared to traditional action recognition. The \textbf{first} is the introduction of background videos that share similar scenes and objects as positive videos but do not contain the target actions. This downplays the role of appearance and static information since background videos are distinguishable from action videos primarily based on the motion. Associated with this is the \textbf{second} difference where the classification task is changed from a forced-choice multi-class formulation to a multi-label \textit{binary} task, where each video can contain multiple actions. This has been enabled through the use of background videos and is not possible with other action datasets. And \textbf{third} is the introduction of untrimmed videos (Figure~\ref{figTeaser}) for validation and testing as opposed to manually pre-segmented (or ``trimmed'') videos~\cite{Schuldt04,Blank05,Rodriguez08,Kuehne11,Soomro12,Liu09} typically used in action recognition. Consequently, a testing video in THUMOS'15 can contain zero, one or multiple instances of an action (or different actions) that can occur anywhere in the given video.


One of the contributions of this paper is to extend and complement prior work with a study of action recognition in temporally untrimmed videos and show how it differs from trimmed videos using the THUMOS dataset (see Fig.~\ref{figTeaser}). We address both video-level action classification and temporal detection problems and systematically evaluate and quantify the effect of temporal clutter. In particular, we evaluate the popular Improved Dense Trajectory Features (IDTF)~\cite{Wang13} + Fisher Vectors + SVM pipeline
that has dominated several action recognition benchmarks. While temporal clutter causes a drop in recognition performance, untrimmed videos also contain additional information about the context of actions. In the evaluation study,  we explore action context and show improvements in action recognition performance using context information extracted from temporal neighborhoods of untrimmed videos.


The rest of the paper is organized as follows. We provide comparison with existing datasets in Sec.~\ref{sec:existing_datasets} and define challenge tasks in Sec.~\ref{sec:tasks}. Next, we explain the procedure used for collection and annotation of the dataset in Sec.~\ref{sec:dataset}, and present the evaluation protocol in Sec.~\ref{sec:sub_eval}. Since the challenge in still nascent, a longitudinal study of participants' methods would be possible after the next few years. Nonetheless, we perform a cross-sectional study of the THUMOS'15 challenge with a summary of methods presented in Sec.~\ref{sec:methods} and results reported in Sec.~\ref{sec:results}. Additionally, we study the impact of background and temporal clutter, as well as role of context for action recognition in untrimmed videos in Sec.~\ref{sec:context}. Finally, we conclude with ideas on improvements for future challenges in Sec.~\ref{sec:future}.

\section{Related Datasets}\label{sec:existing_datasets}

Early datasets on action recognition in video, such as \textbf{KTH}~\cite{Schuldt04} and \textbf{Weizmann}~\cite{Blank05}, employed actors performing a small set of scripted actions under controlled conditions. The next series of datasets, such as \textbf{CMU}~\cite{Ke05} and \textbf{MSR Actions}~\cite{Yuan09}, introduced scripted actions performed against challenging dynamic backgrounds. Later datasets, such as \textbf{HOHA}~\cite{Laptev08} and \textbf{Hollywood-2}~\cite{Marszalek09} 
moved to relatively more realistic video footage from Hollywood movies and broadcast television channels, respectively. Many of these datasets provided spatiotemporal annotations for action instances in relatively short untrimmed videos. However, this level of annotation became impractical once the research community demanded larger datasets. Most of the modern datasets are collected from realistic sources, have more classes and have more temporal clutter. For instance, the \textbf{Human Motion DataBase (HMDB)}~\cite{Kuehne11} dataset released in 2011 contains 51 action categories, each containing at least 101 samples for a total $\sim$6800 action instances.

\begin{figure*}
\begin{center}
   \includegraphics[width=1\linewidth]{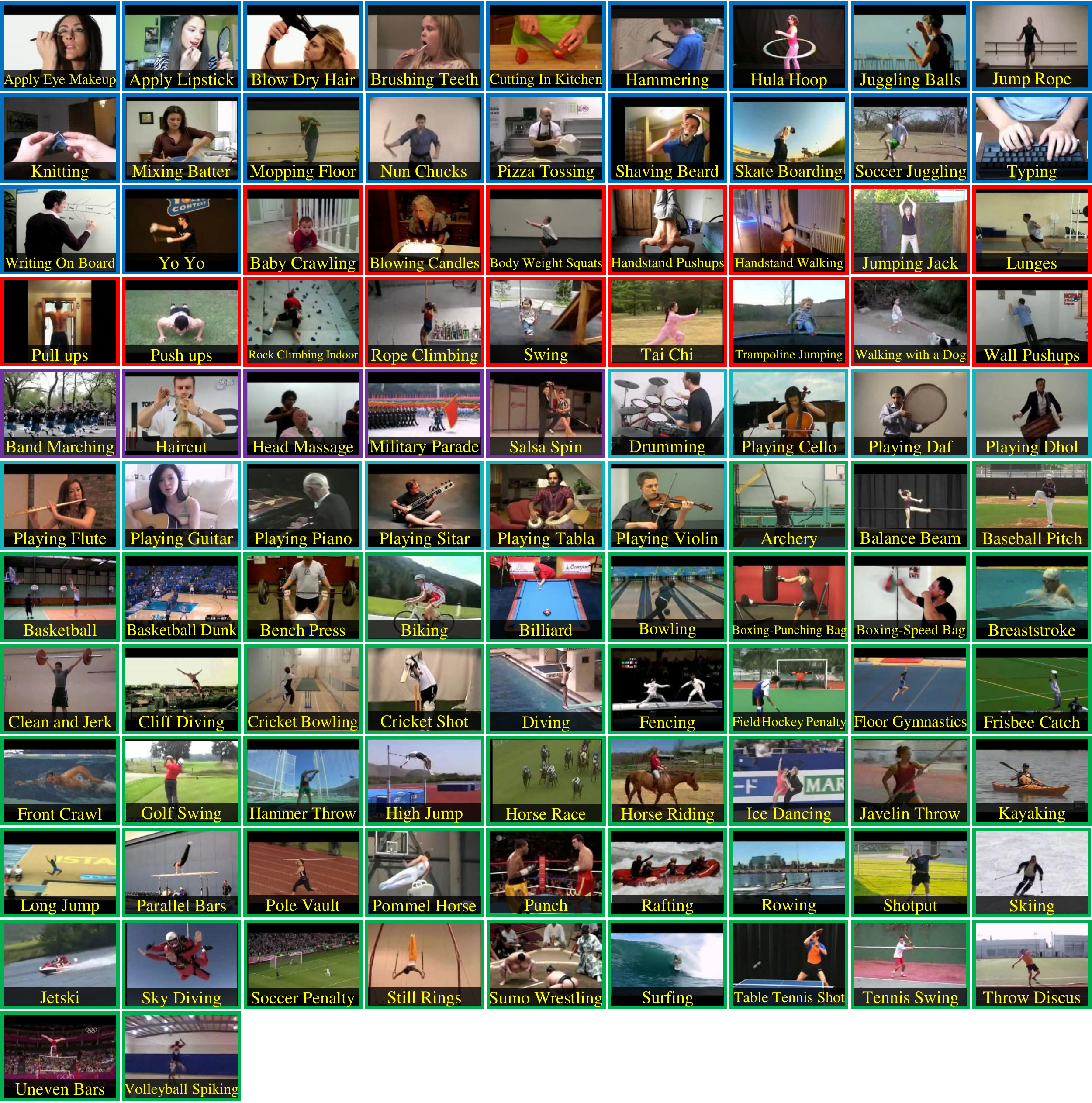}
\end{center}
   \caption{The figure shows the sample frames of the actions from UCF101 dataset~\cite{Soomro12}. The color of frame borders specifies the action type to which they belong: {\color{NavyBlue} Human-Object Interaction}, {\color{red} Body-Motion Only}, {\color{RoyalPurple} Human-Human Interaction}, {\color{YellowOrange} Playing Musical Instruments}, {\color{ForestGreen} Sports} (c.f.\ \ref{sec:appendixA}).}
\label{figUCF101frames}
\end{figure*}

The \textbf{UCF Sports}~\cite{Rodriguez08} dataset from 2009 comprised of movie clips captured by professional filming crew, and similar to many existing datasets at the time, it offered videos with camera motion and dynamic backgrounds. The next in the series were \textbf{UCF11}~\cite{Liu09} and \textbf{UCF50}~\cite{Reddy13}, released in 2009 and 2011, respectively. Both datasets consisted of trimmed clips from a variety of sources ranging from digitized movies to YouTube. The \textbf{UCF101} dataset ~\cite{Soomro12} is a superset of the previous UCF11~\cite{Liu09} and UCF50~\cite{Reddy13} datasets and was released in 2012. It contains 13320 video clips of 101 action classes (\ref{sec:appendixA}). The actions are divided into 5 categories: Human-Object Interaction, Body-Motion Only, Human-Human Interaction, Playing Musical Instruments, Sports, as shown in Figure \ref{figUCF101frames}. The clips of one action class are divided into 25 groups which contain 4--7 clips each. The clips in one group share some common features, such as the background or actors. The videos have a resolution of 320$\times$240, with $\sim$27hrs in total. The training data of the THUMOS challenge uses the trimmed clips of UCF101, however, the datasets for THUMOS'14 and THUMOS'15 additionally include untrimmed positive and background videos for validation and test sets.

The \textbf{Sports-1M}~\cite{Karpathy14} dataset, released in 2014, contains more than 1 million untrimmed videos from almost 487 classes with about 1000--3000 videos per action class. The dataset is divided into the following categories: Aquatic Sports, Team Sports, Winter Sports, Ball Sports, Combat Sports, Sports with Animals, and taxonomy becomes fine-grained at the lower levels. While the dataset is large in the number of videos, it focuses only on sports actions and is weakly annotated (only at the video level) with automatically generated -- and thus potentially noisy -- labels. By contrast, the THUMOS dataset includes videos that have been carefully annotated. Furthermore, THUMOS includes negative background videos for each action class in both the validation and test sets, making the action recognition task more difficult.

\textit{``TREC\footnote{TREC stands for ``Text REtrieval Conference''} Video Retrieval Evaluation'' (TRECVID)}
is a series of competitions and workshops conducted by National Institute of Standards and Technology (NIST) with the aim to stimulate research in automatic segmentation, indexing, and content-based retrieval of digital video. Since the first competition in 2003, it now consists of several independent tasks. The dataset for each task has been typically extended each year, and is only available to the participants who register for the competition. There are two set of tasks in TRECVID that are related to THUMOS challenge. One of the task is \textit{Semantic Indexing} (SIN) and the associated \textit{Localization} (LOC) which focus on the detection and localization in video shots or clips. The dataset consists of \textbf{Internet Archive Creative Commons (IACC)} collected by NIST with 15300 videos for a total of $\sim$1200 hours. Only short clips or shots are annotated for 500 object, scene and action concepts for training. During testing, the highest scoring shots from all participants are gathered, and used for generating ground truth. Since only a subset of test data is annotated, inferred Average Precision is used for evaluation (infAP) \cite{yilmaz2006estimating} of each concept. For 2015, only 30 concepts were evaluated for detection and 10 for spatio-temporal localization. It is important to remember that unlike untrimmed videos in THUMOS, the spatio-temporal localization in SIN task is performed on pre-defined trimmed shots.

Another task \textit{Multimedia Event Detection} requires the methods to provide a confidence score for each video from a collection as to whether the video contains the event. The collection is complemented with event kits that include a textual description of the event and information about related concepts that are likely to occur in each event. An associated task \textit{Multimedia Event Recounting} has the objective of stating key evidence, in the form of text with pointers to detected concepts, that led a Multimedia Event Detection (MED) method to decide that a multimedia clip contains an instance of a specific event. There were 20 pre-specified events for the main task, and Mean Average Precision and inferred MAP used as metrics for event detection. The evaluation for recounting is performed after results are returned by participants where judges evaluate the key evidences for correctness. The dataset consists of \textbf{Heterogeneous Audio Visual Internet (HAVIC)} Corpus collected by the Linguistic Data Consortium. For 40 events, it has $\sim$290 hrs of training videos. The testing is performed on a separate set with 200,000 videos ($\sim$8000 hrs). THUMOS challenge focuses on actions, which are less complex and more atomic than events, and are primarily affected by motion of actors. Furthermore, the action concepts in the Multimedia Event Recounting task are primarily driven by events rather than the actions themselves. Thus, miss-detections of actions are not penalized in evaluation as long as the evidence presented by a system is sufficient for detection of an event.

\textbf{ActivityNet} is a recent dataset for recognition of human activities. It was released in 2015, two years after THUMOS, and consists of 203 activity classes with an average of 137 untrimmed videos per class. The classes are linked through a taxonomy consisting of parent-child relationships. Different from ActivityNet, THUMOS contains a large number of background videos making the problem of action recognition more realistic. For training the classifiers, the negative videos not only come from positive samples of other actions but the background videos associated with an action as well. Thus, it becomes crucial for the classifier and detector to accurately model the motion since similarity in scene in action and background videos significantly reduces the utility of appearance features. The background videos in THUMOS also aid in studying and quantifying the role of stationary and non-action context for action recognition (Sec.~\ref{sec:context}).


\section{The THUMOS Challenge Tasks}\label{sec:tasks}
This section gives an overview of the THUMOS \textit{classification} and \textit{temporal detection} tasks. We also describe their evolution since the first THUMOS held in 2013.

\subsection{Classification}
The task of action classification consists of predicting (for each video) the presence or absence of each of the 101 action classes from the UCF101 dataset. This is a \textit{binary classification} task per action, as the actions are not mutually exclusive --- a given action may occur once, multiple times or never in a testing video. This is in contrast to the typical forced-choice multi-class task whose goal is to assign a class label to a given video from a set of pre-defined classes. For the classification task, the participants are expected to provide real-valued confidences for each test video for all the 101 actions. A low confidence for a particular action means either the video contains some other action or none of the 101 actions. The participants are required to report results on all the videos, and omitting videos from evaluation results in lower performance.

The classification task of the 2013 challenge only consisted of videos from UCF101. The dataset was divided into three pre-defined splits and participants reported results using three-fold cross-validation, i.e., training on two folds and testing on the third. However, since 2014 the dataset has been extended with untrimmed validation, background and test videos. The participants can only use UCF101, validation and background sets to train, validate and fine-tune their models and then report results on the withheld test set. Participants are not permitted to perform any manual annotation at their end.

\subsection{Temporal Detection}

For the temporal detection task participants are expected to provide temporal intervals and corresponding confidence values for all detected instances of 20 pre-selected action classes.
The task of classification is embedded within the temporal detection which makes it comparatively more difficult. For example, an instance of an action that is correctly localized in time but is assigned with an incorrect class label will be treated as an incorrect detection. For this task, participants are required to report results for 20 action classes in all the test videos. For the detection tasks, similar to classification, participants are not permitted to perform additional manual annotations. 

The first THUMOS challenge in 2013 had spatio-temporal localization for 24 action categories instead of temporal detection. The spatio-temporal annotations for 24 actions were provided in the trimmed videos of UCF101. The temporal detection resembles spatio-temporal localization with the difference that the spatial location of the detections is not incorporated in the evaluation. Besides the significant reduction in annotation effort, adopting temporal detection over spatio-temporal localization in later years of the THUMOS challenge was driven by two factors. First, temporal detection is computationally more tractable, particularly in long untrimmed videos. Second, in many practical scenarios, the temporal aspect is more important than the spatial, e.g., a user may want to seek directly to the portion of the video that includes the given action and may not benefit from a bounding box localizing the action within each frame. For these reasons, the 2014 and 2015 challenges only included a temporal detection task, with both the training and test set containing temporal annotations in untrimmed videos for the 20 actions.


\section{The THUMOS Dataset}\label{sec:dataset}

This section provides an overview of the data collection and annotation procedures. In addition, we also provide various statistics related to the THUMOS'15 dataset.

\subsection{Video Collection Procedure}
The Internet videos for the THUMOS competitions were drawn from public videos on YouTube, which made it possible to find a large number of videos for any given topic --- but a large fraction of videos may not contain visible instances of the desired action. We employed a series of manual filtering stages to ensure the set of videos for each action contains only the relevant videos.

\bigskip\textbf{Positive Videos:} The YouTube Data API\footnote{\url{https://developers.google.com/youtube/v3/}} allows video search through Freebase\footnote{\url{https://developers.google.com/youtube/v3/guides/searching_by_topic}} topics. Every YouTube video has several Freebase topics associated with it that are assigned based on annotations provided by the video creator, as well based on some high level video features. We defined a set of Freebase topics corresponding to the action labels. However, a Freebase topic which ideally corresponds to an action either returns too few videos or is too general to be useful. Therefore, we manually augmented topic ids with a set of search keywords. Keywords combined with Freebase topics yielded a reasonable set of potential videos for each action.

An issue with YouTube videos in context of our task is that highly rated or frequently viewed videos may include ``viral'' videos or compilations, so we had to exclude these by explicitly blacklisting keywords ``-awesome", ``-crazy", ``-compilation", etc. Furthermore, as the dataset is extended each year by collecting new videos, we exclude all YouTube videos and channels whose videos were used in previous THUMOS competitions to avoid adding videos that might be similar to those from previous years.

\bigskip\textbf{Background Videos:} Collecting \textit{useful} background videos is more involved than searching for positive videos. Simply adding videos from unrelated categories does not help since such videos are visually dissimilar to those in the positive set. The best background videos are those that share the \emph{context} of a given action (i.e., include similar scenes, actors and objects) without actually showing instances of the given action being performed. For instance, for the `PlayingPiano' class, a video showing a piano in which the piano is not being played is a valid background video. It is also important that background videos for one action class do not contain positive instances of other actions. Therefore, for this task we grouped all action types into super classes. Several actions occur in similar settings: e.g., `BalanceBeam', `FloorGymnastics', `ParallelBars', etc. are all likely to occur indoors in Olympic gymnastic venues; whereas `HammerThrow', `HighJump', `HighJump', etc., occur outdoors in track and field arenas. To find such videos, we supplemented the search with the following queries which resulted in background videos without any instance of that action:

\bigskip
\begin{itemize}[noitemsep,nolistsep]
{
\item \textbf{X + `for sale'}: for actions that involve an instrument, e.g., piano for sale (`PlayingPiano'), yoyo for sale (`YoYo').
\item \textbf{X + venue}: for actions that involve a particular location or venue, e.g. baseball stadium or Coors Field (`BaseballPitch'), climbing tower (`RockClimbing'), bathroom (`BrushingTeeth').
\item \textbf{Co-occurring events}: for sports related actions, e.g., cheer leading or dance, e.g., waist twirling dance -hoop -contra (`HulaHoop').
\item \textbf{X + brands}: for actions involving branded objects e.g., L'oreal eye makeup (`ApplyEyeMakeup').
\item \textbf{X + `drill' or `workout'}: for some sports actions, e.g., shotput drill (`ShotPut').
\item \textbf{X + `review' or `how to choose'}: for products, e.g., lipstick overview (`ApplyLipstick').
\item \textbf{General Freebase topics}: excluding class names e.g., circus gymnastics (`StillRings'), computer (`Typing'), macram\'{e} (`Knitting').
\item \textbf{Object names}: for actions involving object e.g., `piano -playing' (`PlayingPiano'), bat (`CricketShot').
\item \textbf{Different object / action combination}: mechanical bull ride (`PommelHorse'), Invisible drum (`PlayingTabla'), running with dog (`WalkingDog'), yoga standing pose (`Lunges').
}
\end{itemize}
\bigskip

The video collection procedure builds lists of putative positive and background videos for each action class. The \textit{YouTube id}, \textit{channel id}, and \textit{title} of each video are saved in the list. Next, the videos go through an annotation stage, followed by 
downloading and final verification.

\subsection{Annotation and Verification Procedure}
The video collection procedure provides a set of potential positive and background videos for each of the 101 action classes. For positive videos, the annotators were asked to first go through the videos of a particular action class in UCF101, and then annotate the videos from the list as either \emph{positive} or \emph{irrelevant}. The videos for a particular action were presented to the annotator in a batch of four (for User Interface efficiency reasons), which were played simultaneously from YouTube. As soon as the annotator found a positive and valid instance of the action class being annotated, s/he marked it as positive. A video may contain an instance of an action, but was marked as \emph{irrelevant} if it satisfied any of the following criteria:

\bigskip
\begin{itemize}[noitemsep,nolistsep]
\item \textbf{Slow Motion}: The video contains action that has been performed in slow motion or in an unrealistic way, and looks different from the instances of an action class in UCF101 dataset.
\item \textbf{Sped Up}: The action is being performed faster than usual.
\item \textbf{Occlusions / Partial Visibility}: There is text or any other object significantly occluding the actor.
\item \textbf{Motion Blur}: Video is blurry or camera is shaking to the extent that the action cannot be seen properly.
\item \textbf{Clutter / Incorrect Background}: Action is performed in an environment where it is partially visible e.g., a `GolfSwing' action recorded from a camera directly behind the audience, therefore they are blocking the field-of-view, or if it has an atypical backdrop, e.g., somebody performing `PushUps' on the moon.
\item \textbf{Unrealistic Instances}: The action does not seem realistic. For example, an instructional video on how to perform a `PushUp' might have a person performing the action much slower than usual. The person might also stop half-way while performing the action to explain, or performs an action in an unusual way, not seen in the UCF101 dataset.
\item \textbf{Animation}: Any animated examples of the action of interest, e.g. a character from a video game performing the action or from a cartoon, etc.
\item \textbf{Fake Action}:	The action does not seem realistic or is poorly performed.
\item \textbf{Long Video}: Video is longer than 10 minutes.
\item \textbf{Compilation}:	Video is compiled using multiple videos.
\item \textbf{Slide Show of Images}: The video contains a slide show of images, but no video of the action of interest.
\item \textbf{First Person Video}: The video is recorded from an egocentric perspective by the same person who is performing the action i.e. actions viewed from a wearable camera.
\item \textbf{Not Related}:	The video neither contains any instance of the action of interest nor the background for that action.
\end{itemize}
\bigskip

The positive videos are also annotated with secondary actions, ones which occur or co-occur with the primary action in a video. Some of the actions are subset of others, for instance, `BasketballDunk' implies `Basketball', `HorseRace' implies `HorseRiding', and `CliffDiving' implies `Diving'. Similarly, there are several actions that are usually proximal in time, such as `CricketBowling' and `CricketShot', as well as videos involving playing of musical instruments that can have multiple secondary actions. In contrast to positive videos, the task of annotating background videos is somewhat more difficult as each background should not contain instances of any of the 101 action classes. To achieve this, each annotator was asked to review at most 34 actions at a time, and ensure none of those occurred in the background video being annotated. Thus, each background video was annotated by three different annotators for three distinct subsets of 101 action classes. Once the annotation is finished for positive and background videos, all of them are verified by a different set of annotators both for consistency and accuracy.

\subsection{Temporal Annotations}

Action boundaries (unlike objects) are generally vague and subjective. This makes the evaluation less concrete as human experts define the action boundaries differently from each other. The same is true for different methods whose output can vary among each other. However, we observed that the 101 action classes can be divided into two categories: the \textit{instantaneous} actions which have short time span and can be well-localized in time e.g., `BasketballDunk', `GolfSwing'; and \textit{cyclic} actions that are repetitive in nature, e.g. `Biking', `HairCut', `PlayingGuitar'. To select the action classes for the temporal detection task, we handpicked the instantaneous ones\footnote{\label{foot:localizationclasses} BaseballPitch (07), BasketballDunk (09), Billiards (12), CleanAndJerk (21), CliffDiving (22), CricketBowling (23), CricketShot (24), Diving (26), FrisbeeCatch (31), GolfSwing (33), HammerThrow (36), HighJump (40), JavelinThrow (45), LongJump (51), PoleVault (68), Shotput (79), SoccerPenalty (85), TennisSwing (92), ThrowDiscus (93), VolleyballSpiking (97).} with well-defined temporal boundaries (c.f.\ \ref{sec:appendixA}).

Besides only focusing on instantaneous actions for the temporal detection, we also take additional measures to ensure that evaluation for this task is objective. First, we annotated action intervals consistently with the temporal segmentation of corresponding actions in the UCF101 dataset. Second, we also marked some action instances as ambiguous in cases of partial visibility, incomplete execution or strong deviation in the style. Third, we use a liberal Intersection-Over-Union threshold (small, 10\%) to quantify the performance on this task, since actual actions are only a small fraction of the entire videos. Lastly, we ensured that evaluation at multiple IOU thresholds keeps the rankings unaffected.

\begin{figure*}[t]
\centering
\includegraphics[width=1\linewidth]{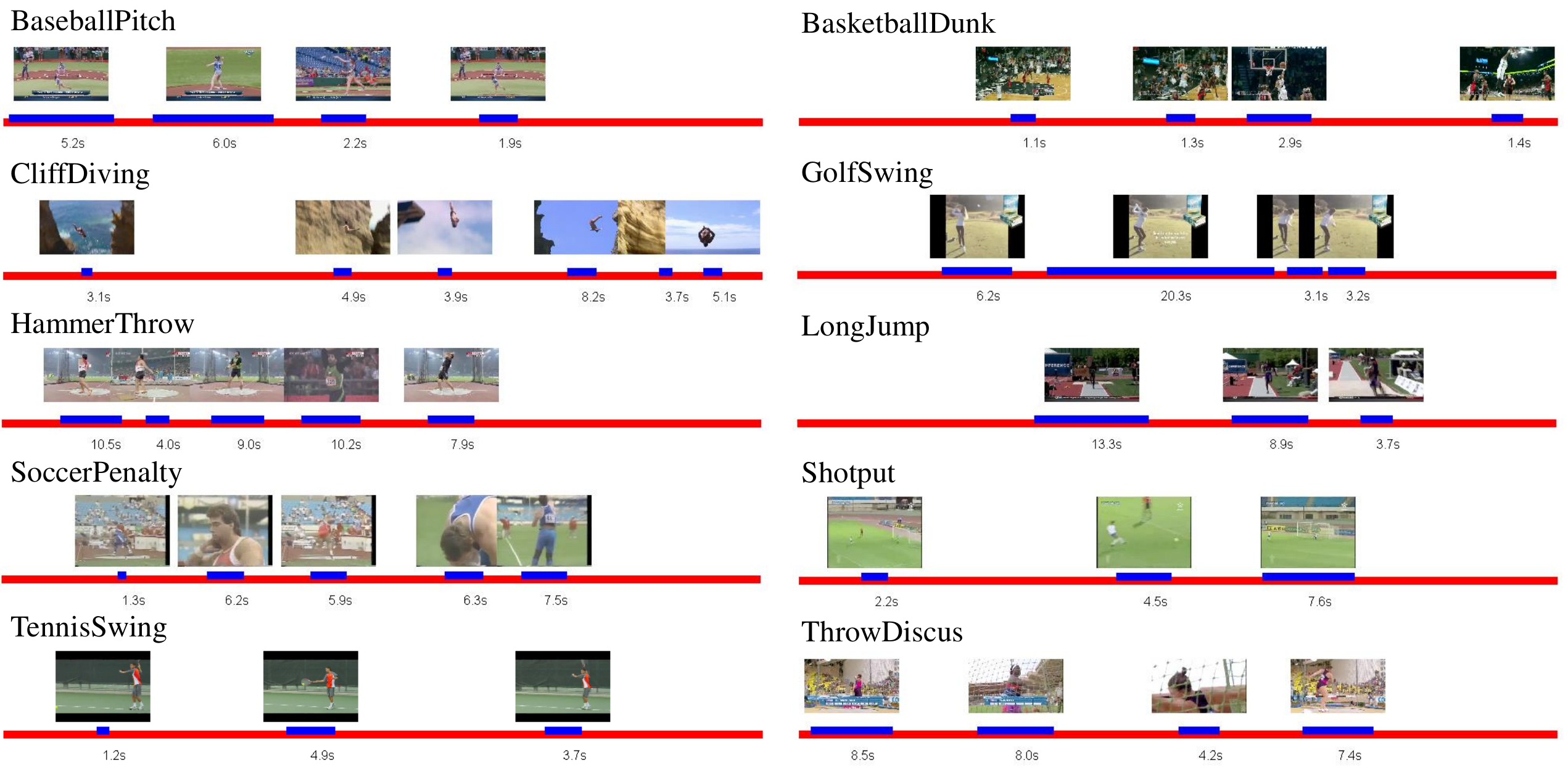}
\caption{Illustration of temporal annotation (shown in blue) for eight video samples from the Validation set of THUMOS'15 dataset.}
\label{figTemporalAnnotation}
\end{figure*}

For the 20 instantaneous actions selected for the task of temporal detection, we annotated their temporal boundaries in untrimmed videos. Each instance of these action classes is annotated with the start and end time in all videos in the Validation and Test sets. The labels include any of the 20 actions or \emph{`ambiguous'}. To ensure consistency, the annotation has been made by one annotator in two passes over the data, and then verified by another annotator. The annotation has been performed using the Viper\footnote{\url{http://viper-toolkit.sourceforge.net/products/gt/}} tool. Action annotation for a few example videos is illustrated in Figure~{\ref{figTemporalAnnotation}}. In these and other examples each video typically contains instances of one action category only. Exceptions include `CricketBowling' and `CricketShot' actions which often co-occur within the same video.

\subsection{Attributes}
Besides the video and clip level annotations provided with the THUMOS dataset, we also provided semantic relationships between the 101 action classes and several attributes. Each action class is associated with one or more of these attributes, as summarized in Table~\ref{tableAttributes}. Although video-level annotations for the attributes are not provided, such semantic knowledge can be incorporated while training and testing action categories.

\begin{table*}[t]
\centering
\includegraphics[width=1\linewidth]{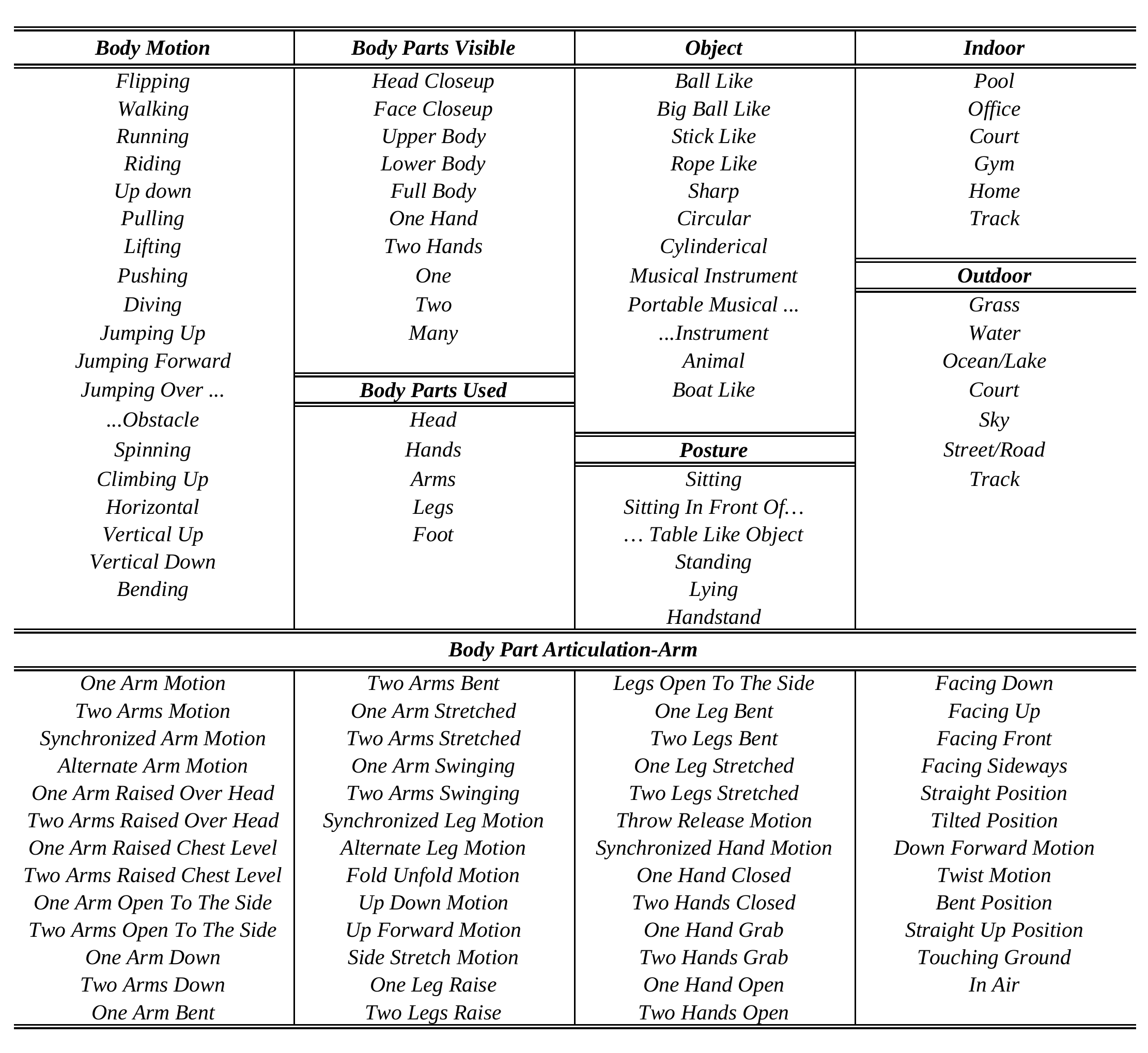}
\caption{Attributes for the 101 action classes.}
\label{tableAttributes}
\end{table*}

\subsection{Dataset Statistics}

We summarize the statistics of THUMOS'15 benchmark dataset below:

\begin{itemize}
\item Validation set: 2,104 untrimmed videos with temporal annotations of actions. This set contains on average 20 videos for each of the 101 classes found in the UCF101 dataset.
\item Background set: 2,980 relevant videos that are guaranteed not to contain any instances of the 101 actions.
\item Test set: 5,613 untrimmed videos with temporal annotations for 20 classes.
\end{itemize}

\begin{figure}[t]
\centering
\includegraphics[width=.7\linewidth]{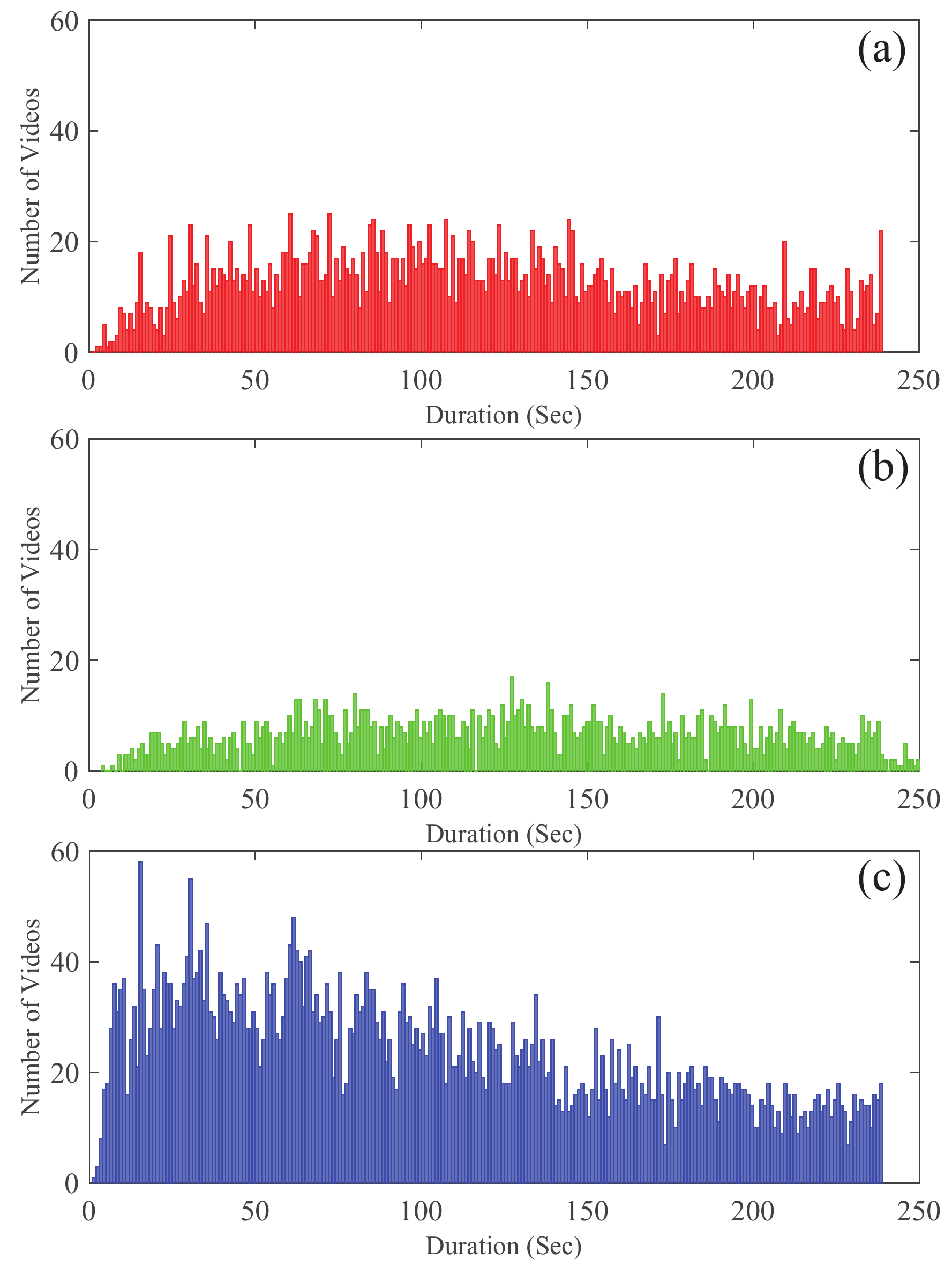}\\
\caption{Histogram of video lengths in THUMOS'15 (a) Background, (b) Validation and (c) Test set, respectively. We excluded videos from the Validation set which were over 250 seconds long.}
\label{figStats}
\end{figure}

The THUMOS'15, which is an extension of THUMOS'14 dataset, was designed to provide a realistic action recognition scenario. Unlike UCF101~\cite{Soomro12}, the videos in the set were not temporally segmented to contain only the actions of interest. Therefore, in most of the videos the action only takes a small percentage of time when compared to the length of the video in which it occurs (see Fig.~\ref{figStats}) (the only notable exceptions are videos of cyclic actions).  The use of variable length videos, each containing different numbers of actions of different lengths makes it less likely that a system could inadvertently exploit side-information~\cite{Satkin10}, such as action length during the classification task. The mean clip length for UCF101 is 7.21 seconds, which is about 80\% more than the average action length in the THUMOS'15 dataset.

\begin{table*}[t]
\scalebox{0.8}{
\footnotesize
\centering
\tabcolsep=0.07cm
\begin{tabular}{l||c c c c c c c c c c c c c c c c c c c c c}
\hline
\hline
\textbf{Action ID}           & \textbf{07}   & \textbf{09}   & \textbf{12}   & \textbf{21}    & \textbf{22}   & \textbf{23}   & \textbf{24}   & \textbf{26}   & \textbf{31}   & \textbf{33}   & \textbf{36}   & \textbf{40}   & \textbf{45}   & \textbf{51}   & \textbf{68}   & \textbf{79}   & \textbf{85}   & \textbf{92}   & \textbf{93}   & \textbf{97}   & \textbf{All} \\
\hline
\hline
Instances           & 71   & 791  & 187  & 140   & 360  & 316  & 351  & 887  & 151  & 67   & 441  & 406  & 361  & 305  & 519  & 214  & 114  & 210  & 208  & 266  & 6365 \\

Length (secs)       & 3.1  & 1.8  & 2.7  & 11.9  & 3.1  & 1.7  & 1.4  & 3.1  & 3.0  & 8.6  & 7.5  & 5.1  & 6.6  & 7.4  & 7.2  & 5.5  & 3.2  & 2.2  & 5.0  & 2.6  & 4.6\\

Ratio               & 12.2 & 24.0 & 14.5 & 47.8  & 27.1 & 13.5 & 11.6 & 29.7 & 38.2 & 30.3 & 40.8 & 31.3 & 24.6 & 31.5 & 40.2 & 33.7 & 19.2 & 22.6 & 39.3 & 28.5 & 28.0 \\
\hline
\hline
\end{tabular}}
\caption{Statistics of temporal annotation for 20 action classes in the validation set of THUMOS'15 dataset. For each class the rows indicate (i) the number of actions instances, (ii) the average length of action intervals in seconds and (iii) the ratio of action length with respect to the length of the video.}
\label{tab:stats}
\end{table*}

Statistics of the temporal annotation for the 20 action classes in the Validation set is presented in Table~\ref{tab:stats}. As can be seen, the average length of such actions is $\sim$4.6~seconds while their temporal intervals occupy $\sim$28\% of corresponding videos. The relatively large number of action instances and the low ratio of action length indicate the difficulty of the THUMOS temporal detection task.

\section{Submission and Evaluation}\label{sec:sub_eval}
\subsection{Action Recognition}
For action recognition, each system is expected to output a real-valued score indicating the confidence of the predicted presence in a video. Due to the untrimmed nature of the videos, a significant part of a test video may not include any particular action, and multiple instances may occur at different time-stamps within the video. Similarly, the video may not contain any of the actions, for which the expected confidence for each action is zero.

Each team was allowed to submit the results of at most five runs. The run with the best performance is selected as the primary run of the submission and is used to rank the teams. Each run has to be saved in a separate text file with 102 columns\footnote{Sample output for Classification: \url{http://goo.gl/sNQQBh}}, where the first column contains the name of the test video, and rest of the columns contain confidences for the 101 actions. Essentially, each row shows the results of one test video, and each column contains the confidence score of presence of the corresponding action class anywhere in the video. The confidence scores must be between 0 and 1.  A larger confidence value indicates greater confidence to detect the action of interest in a test video.

We use \textbf{Interpolated Average Precision (AP)} or \textbf{11-Point Average Precision} as the official measure for evaluating the results on each action class. Given a descending-score-rank of videos for the test action class c, the AP(c) is computed as:
\begin{equation}
AP(c) =  \frac{\sum_{k=1}^{n}(\textrm{Prec}(k) \times \textrm{rel}(k))}{\sum_{k=1}^{n}\textrm{rel}(k)},
\end{equation}
where $n$ is the total number videos, $\textrm{Prec}(k)$ is the precision at cut-off $k$ of the list, $\textrm{rel}(k)$ is an indicator function equaling to $1$ if the video ranked $k$ is a true positive, and to zero otherwise. The denominator is the total number of true positives in the list. Mean Average Precision (mAP) is then used to evaluate the performance of one run over all action classes.

\subsection{Temporal Detection}

Temporal detection is evaluated for twenty classes of instantaneous actions\textsuperscript{\ref{foot:localizationclasses}} in all test videos. The system is expected to output a real-valued score indicating the confidence of the prediction, as well as the starting and ending time for the given action\footnote{Sample output for Temporal Detection: \url{http://goo.gl/SWZbBM}}. For this task, each team is allowed to submit at most 5 runs. The run with the best performance is selected as the primary run of the submission and is used to rank across teams. Each run must be saved in a separate text file with the following format, where each row represents one detection output by the system:
\begin{equation}\notag
\textrm{[video name]  [starting time] [ending time] [class label] [confidence score]}
\end{equation}

Each row has five fields representing a single detection. A detector can fire multiple times in a test video (reported using multiple rows in the submission file). The time must be in seconds with one decimal point precision. The confidence score should be between 0 and 1.

For evaluation, detected time intervals of a given class are sorted in the order of decreasing detector confidence and matched to ground truth intervals using Intersection over Union (IoU, also known as Jaccard) similarity measure. Detections with IoU above a given threshold are declared as true positives. To penalize multiple detections of the same action, at most one detection is assigned to each annotated action and the remaining detections are declared as false positives. Annotations with no matching detections are declared as false negatives. Given labels and confidence values for detections, the detector performance for an action class is evaluated by Average Precision (AP). The mean AP value for twenty action classes (mAP) provides the final performance measure for a method. To account for somewhat subjective definition of action boundaries, the evaluation is reported for different values of IoU threshold (10\%, 20\%, 30\%, 40\%, and 50\%). Action intervals marked as ambiguous are excluded from the evaluation, hence, all detections having non-zero overlap with ambiguous intervals are ignored.

\section{Methods}\label{sec:methods}
This section presents methods used by participants for both tasks at the THUMOS'15 challenge. A comprehensive survey of techniques and their evolution across years is beyond the scope of this paper, and will be made after several more challenges in the future.

\subsection{Classification}
In this subsection we briefly summarize the classification methods of the 11 teams. Table~\ref{tb:clsMethod} summarizes the major feature extraction and fusion methods. Most teams adopted two kinds of features, deep learning based features and the improved Dense Trajectories (iDT) \cite{Wang13}.

\renewcommand{\multirowsetup}{\centering}
\begin{table*}[th!]
 \centering
 \small
 \setlength{\tabcolsep}{5pt}
 \begin{tabular}{c||cccccc|ccc|cccc}
  \hline
  \hline
  \multirow{8}{3.0cm}{\bf{Team}} & \multicolumn{6}{c|}{\multirow{2}{3.2cm}{\bf{Deep Features: ~~~~~~~ Structures \& Encoding}}} & \multicolumn{3}{c|}{\multirow{2}{1.5cm}{\bf{Traditional Features}}} &\multicolumn{4}{c}{\multirow{2}{1.5cm}{\bf{Fusion Methods}}} \\
  &&&&&&&&&&&&& \\
  \cline{2-14}
  &\rotatebox{90}{VGGNet} & \rotatebox{90}{GoogleNet} &\rotatebox{90}{FC 6,7,8} & \rotatebox{90}{LCD} &\rotatebox{90}{VLAD} & \rotatebox{90}{Mean/Max Pool} & \rotatebox{90}{iDT} & \rotatebox{90}{MFCC} & \rotatebox{90}{ASR} & \rotatebox{90}{Average} &\rotatebox{90}{Logistic Regression~} &\rotatebox{90}{Weighted} &\rotatebox{90}{Geometric Mean}\\
  \hline
  \hline
  \multirow{1}{3.0cm}{UTS \& CMU \cite{xu2015uts}}  &$\bullet$ &$\bullet$	 &$\bullet$&$\bullet$	&$\bullet$ &--		 &$\bullet$&$\bullet$&$\bullet$	 &--&$\bullet$&--&--\\
  \multirow{1}{3.0cm}{MSR Asia (MSM) \cite{qiumsr}} &$\bullet$ &--	 &$\bullet$&--		&-- &$\bullet$		 &$\bullet$&$\bullet$&--	&$\bullet$&--&--&--\\
  \multirow{1}{3.0cm}{Zhejiang U. \cite{wuzjudcd}}  &$\bullet$ &--		 &$\bullet$&$\bullet$	&$\bullet$ &--		&$\bullet$&--&-- 	 &$\bullet$&--&--&--\\
  \multirow{1}{3.0cm}{INRIA LEAR \cite{peng2015encoding}} &$\bullet$ &--		 &$\bullet$&$\bullet$	&$\bullet$ &$\bullet$	 &$\bullet$&--&--	 &$\bullet$&--&--&--\\
   \multirow{1}{3.0cm}{CUHK \& SIAT \cite{wang2015cuhk}} &$\bullet$ &$\bullet$	 &$\bullet$&--	&-- &$\bullet$		&$\bullet$&--&--	 &$\bullet$&--&--&--\\
  \multirow{1}{3.0cm}{U. Amsterdam \cite{jainuniversity}}  & -- &$\bullet$		 &$\bullet$&--		&$\bullet$&--		&$\bullet$&--&--	 &$\bullet$&--&--&--\\
  \multirow{1}{3.0cm}{Tianjin U. \cite{liutianjin}}  &$\bullet$&--			 &--&$\bullet$		&$\bullet$&--		&$\bullet$&--&--	 &--&--&$\bullet$&--\\
  \multirow{1}{3.0cm}{USC \& THU \cite{ganusc}}  &$\bullet$ &--	 &$\bullet$&--		&-- &$\bullet$		&$\bullet$&--&--	 &--&--&--&$\bullet$\\
  \multirow{1}{3.0cm}{U. Tokyo \cite{ohnishimil}} &$\bullet$ &--			 &$\bullet$&--		&$\bullet$ &--		&$\bullet$&--&-- 	 &$\bullet$&--&--&--\\
  \multirow{1}{3.4cm}{ADSC, NUS \& UIUC \cite{yuanadsc}}  &$\bullet$ &--		 &$\bullet$&--		&-- &$\bullet$		&$\bullet$&--&--	 &$\bullet$&--&--&--\\
  \multirow{1}{3.0cm}{UTSA \cite{caiutsa}}  &$\bullet$ &--			 &$\bullet$&--		&-- &$\bullet$		&--&--&--			 &--&--&--&--\\
   \hline
   \hline
\end{tabular}
 \setlength{\tabcolsep}{6pt}
 \caption{The major feature extraction and fusion methods of all the teams. Here, the symbols $\bullet$ and -- represent the presence or absence of a feature or technique, respectively. }
\label{tb:clsMethod}
\end{table*}


Deep learning features extracted by Convolutional Neural Networks (CNN) have been popular in many visual recognition tasks. By considering different network architectures and feature pooling methods, the resulting CNN features may vary greatly. For network architectures, VGGNet~\cite{simonyan2014very}, GoogleNet~\cite{szegedy2014going}, ClarifaiNet~\cite{zeiler2014visualizing} and 3D ConvNets (C3D)~\cite{C3D} were used. In particular, VGGNet was used by most teams, and GoogleNet was used by three teams (UTS\&CMU, CUHK\&SIAT, UvA). Each of the remaining two networks was used by only one team (CUHK\&SIAT used ClarifaiNet, and MSM used C3D), which are therefore excluded from the table due to space limitations. In addition, the recent two-stream CNN approach \cite{Simonyan14}, which explores both spatial stream (static frames) and temporal stream (optical flows), was adopted by the CUHK\&SIAT team.

For the CNN based models, typically the outputs of $6^\text{th}$, $7^\text{th}$ or $8^\text{th}$ fully connected layers (FC6, FC7, FC8) are used as features. A few teams also explored a recent method called latent concept descriptors (LCD) \cite{LCD-work}. In addition, as the CNN features are computed on video frames, a pooling scheme is needed to convert the frame-level feature into a video-level representation. For this, most teams adopted the Vectors of Locally Aggregated Descriptors (VLAD) \cite{jegou2010aggregating} and the conventional mean/max pooling.

The iDT is probably the most powerful hand-crafted feature for video classification. It extracts four kinds of features, i.e., trajectory shape, HOG, HOF and MBH, on the spatial-temporal volumes along the extracted dense trajectories. The features are encoded with the Fisher Vector (FV) \cite{sanchez2013image} to generate a video level representation. The UTS\&CMU team used a variant of iDT, called enhanced iDT \cite{MSFS-work}. The UTS\&CMU and the MSM teams also used auditory features MFCC and ASR.

For classification, all of the teams adopted SVM as the classifier. In addition, the USC\&Tsinghua team adopted kernel ridge regression (KRR) \cite{yu2014informedia} as an alternative classifier. While the classifiers are consistent across the teams, the fusion method varies. As shown in the table, average fusion is the most popular option due to its simplicity and good generalizability, but there are other strategies like weighted fusion, logistic regression fusion, geometric mean fusion, etc.

\subsection{Temporal Detection}
This section summarizes the methods used for temporal detection of actions in testing videos. For the THUMOS'15 challenge, we received 5 runs from only one team. The team consists of researchers from Advanced Digital Sciences Center (ADSC), National University of Singapore (NUS), and University of Illinois Urbana-Champaign (UIUC). The temporal detection task attracted fewer participants compared to the classification task due to its higher computational requirements. Furthermore, temporal detection is a new problem that was introduced recently in THUMOS. With very few research efforts related to temporal detection in the past, we believe it will gain interest of the wider community resulting in increased participation in the future.

The runs from ADSC, NUS and UIUC were obtained using the following pipeline:
First, the Improved Dense Trajectory (iDT)~\cite{Wang13} features are extracted throughout the video. For forming the Gaussian Mixture Model dictionary, only features from UCF101 are used. The video segments as encoded using Improved Fisher Vectors. The FVs were not normalized to maintain additivity of Fisher Vectors. Besides the motion features, scene features were extracted from VGG-19 deep net model~\cite{chatfield2014return}. In particular, features were made from the last 4096-d rectified linear layer.

Since different actions have different lengths, the team used a pyramid of score distributions as features. For each frame, they used nine windows of 10, 20, $\ldots$, 90 frames around it. The hypothesis was that the scores at the correct window length should be highest, and should vary smoothly for neighboring temporal resolutions. Next, the FV in each window are normalized to obtain Improved FV. This yields 9$\times$101 scores, which are concatenated to form a feature vector. The action confidences are then computed using a 21-class SVM (20 actions, 1 background). Afterwards, they use median filtering on output labels for smoothness.


\section{Results}\label{sec:results}
In this section, we present results and analysis of the approaches from the THUMOS'15 challenge presented in the previous section.
\subsection{Classification}

Next, we summarize and discuss the results of the classification task. We received 47 submissions from the 11 teams. Table~\ref{tb:classification} shows the overall results of all the submissions, measured by mAP. The best mAP from each team is highlighted in bold. The teams are sorted based on their highest mAP.

\renewcommand{\multirowsetup}{\centering}
\begin{table*}[th!]
 \centering
 \small
 \begin{tabular}{c||c||c c c c c}
  \hline
  \hline
  \textbf{Rank} & \textbf{Team} & \textbf{Run1} & \textbf{Run2} & \textbf{Run3} & \textbf{Run4} & \textbf{Run5} \\
  \hline
  \hline
  1 & UTS \& CMU \cite{xu2015uts}  & \textbf{0.7384}  &  0.7157 & 0.7011 & 0.6913 & 0.647  \\
  2 &  MSR Asia (MSM) \cite{qiumsr} &  0.6861 & 0.6869 & 0.6878 & 0.6886 & \textbf{0.6897}  \\
  3 & Zhejiang U. \cite{wuzjudcd}  &  \textbf{0.6876} & 0.6643 & 0.6859 & 0.6809 & 0.5625  \\
  4 & INRIA LEAR \cite{peng2015encoding} & \textbf{0.6814} & 0.6811 & 0.5395 & 0.6739 & 0.6793 \\
  5 & CUHK \& SIAT \cite{wang2015cuhk}  &  0.4894 & 0.5746 & \textbf{0.6803} & 0.6576 & 0.6604  \\
  6 & U. Amsterdam \cite{jainuniversity} & \textbf{0.6798}  & NA  &  NA & NA  & NA \\
  7 & Tianjin U. \cite{liutianjin} & \textbf{0.6666} & 0.6551 & 0.6324 & 0.5514 & 0.5357  \\
  8 & USC \& THU \cite{ganusc} & 0.6354 & \textbf{0.6398} & 0.6346 & 0.5639 & 0.6357   \\
  9 &  U. of Tokyo \cite{ohnishimil} &  0.6159 & 0.6172 & \textbf{0.6174} & 0.6087 & 0.4986  \\
  10 & ADSC, NUS \& UIUC \cite{yuanadsc}  & 0.4471 & 0.3451 & 0.4849 & \textbf{0.4869} & 0.3466  \\
  11 & UTSA \cite{caiutsa} & \textbf{0.3981}  &  NA & NA  & NA  & NA \\
   \hline
   \hline
\end{tabular}
 \caption{Classification Results measured by mAP (\%). Each team could submit up to five runs. The teams are sorted based on their highest mAP.}
\label{tb:classification}
\end{table*}

As discussed earlier, most of the approaches adopted two kinds of features: iDT features and deep learning features. IDT features were used by all the top-10 teams, and deep learning features were used by all the teams. Based on the results, we make the following observations: 1) The LCD coding with the VLAD representation \cite{LCD-work} is very effective; 2) fine-tuning the CNN models can bring further improvements; and 3) some specially designed network structures for video analysis are helpful, e.g., the two-stream CNN \cite{Simonyan14}. Furthermore, the results also indicate that multi-modal fusion with audio clues can consistently improve the results.

\subsubsection{Per-action Results}
\begin{figure*}[t]
\begin{center}
\includegraphics[width=1.0\linewidth]{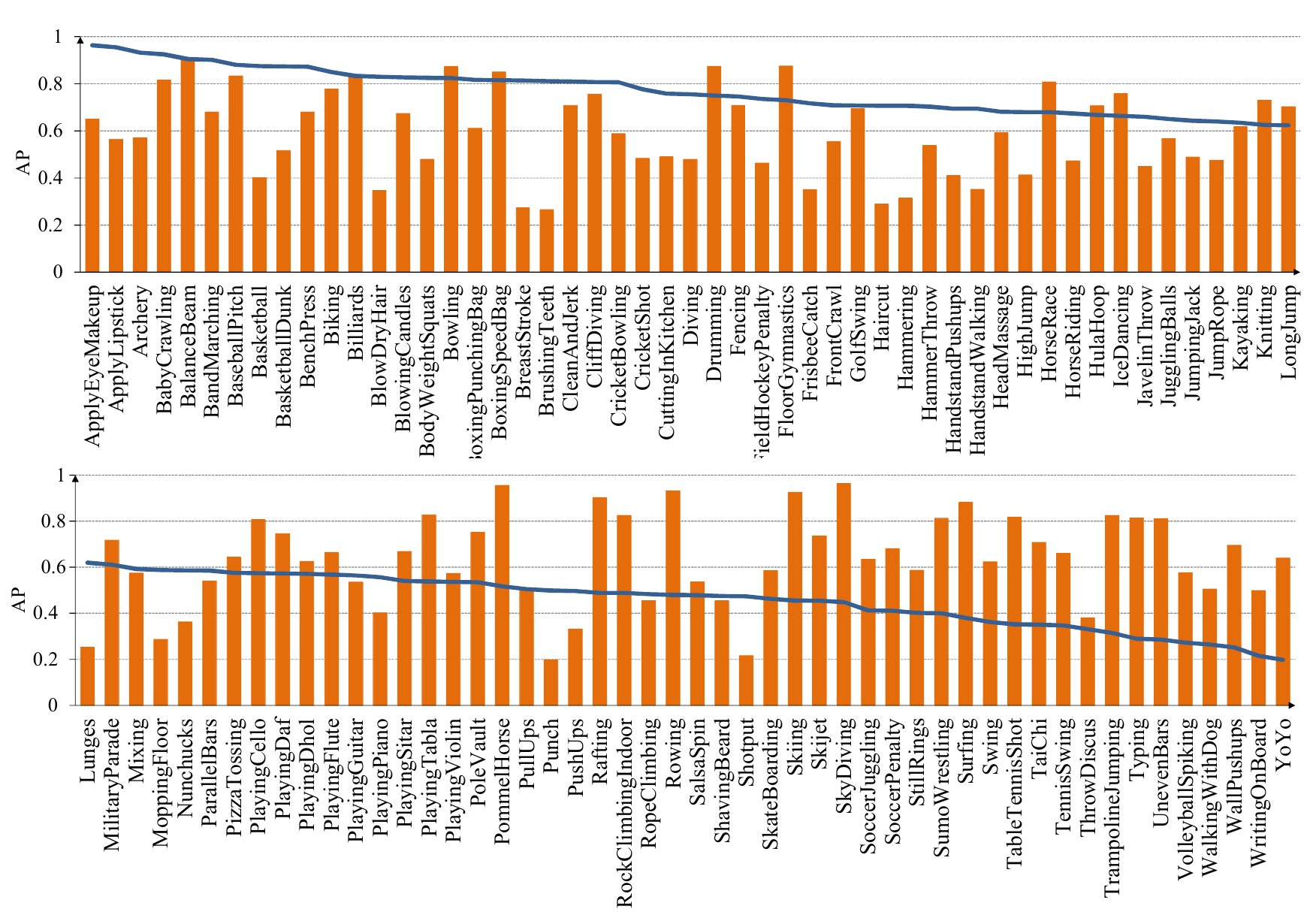}
\end{center}
\caption{Per-action results, measured by AP: The bars depict the AP for each action, and the curve represents the results of all the actions sorted in decreasing AP values. For each action, we report the average AP from all the submissions.}
\label{figPerActionResultClassification}
\end{figure*}

Figure~\ref{figPerActionResultClassification} shows the results of each action class, where the bars depict the AP of each action and the curve represents the results of all the actions sorted in decreasing AP values. For each action, the result is obtained by averaging the results of all the submissions. We can see that the AP varies significantly across different actions, from the lowest value of 19.8\% to the highest of 96.4\%. The curve of sorted AP fits well with a straight line, which indicates that the numbers of actions that are easy/hard to be distinguished are evenly distributed. The mAP over all the action classes is 61.3\%, which reflects an average level of recognition capability of all the teams.

While the results are promising in general, there is still room for improvement. Table~\ref{tb:high_low_class} lists the action classes which are easy or hard to be recognized. Some classes like `Bowling' and `Surfing' are easy but there are many difficult ones that can confuse the classifier. For example, `BlowDryHair' is visually very similar to `Haircut'. More advanced techniques are needed to distinguish these classes.

\renewcommand{\multirowsetup}{\centering}
\begin{table*}[t!]
 \centering
 \small
 \begin{tabular}{c|c||c|c}
  \hline
  \hline
  \textbf{Easy Classes} & \textbf{AP}  & \textbf{Difficult Classes} & \textbf{AP} \\
  \hline
  \hline
  SkyDiving & 0.964  & Punch & 0.198 \\
  PommelHorse & 0.955  & ShotPut & 0.216 \\
  Rowing & 0.933  & Lunges & 0.252 \\
  Skiing & 0.925  & BrushingTeeth & 0.265 \\
  BalanceBeam & 0.905  & BreastStroke & 0.273 \\
  Rafting & 0.902  & MoppingFloor & 0.286 \\
  Surfing & 0.881  & Haircut & 0.290 \\
  FloorGymnastics & 0.875  & Hammering & 0.315 \\
  Drumming & 0.873  & PushUps & 0.331 \\
  Bowling & 0.872  & BlowDryHair & 0.347 \\
  \hline
  \hline
\end{tabular}
 \caption{The top 10 easy and difficult classes in THUMOS'15.}
\label{tb:high_low_class}
\end{table*}

\begin{figure*}[t!]
\begin{center}
   \subfigure[Bowling]{\label{fig:pr_class16}\includegraphics[width=0.44\linewidth]{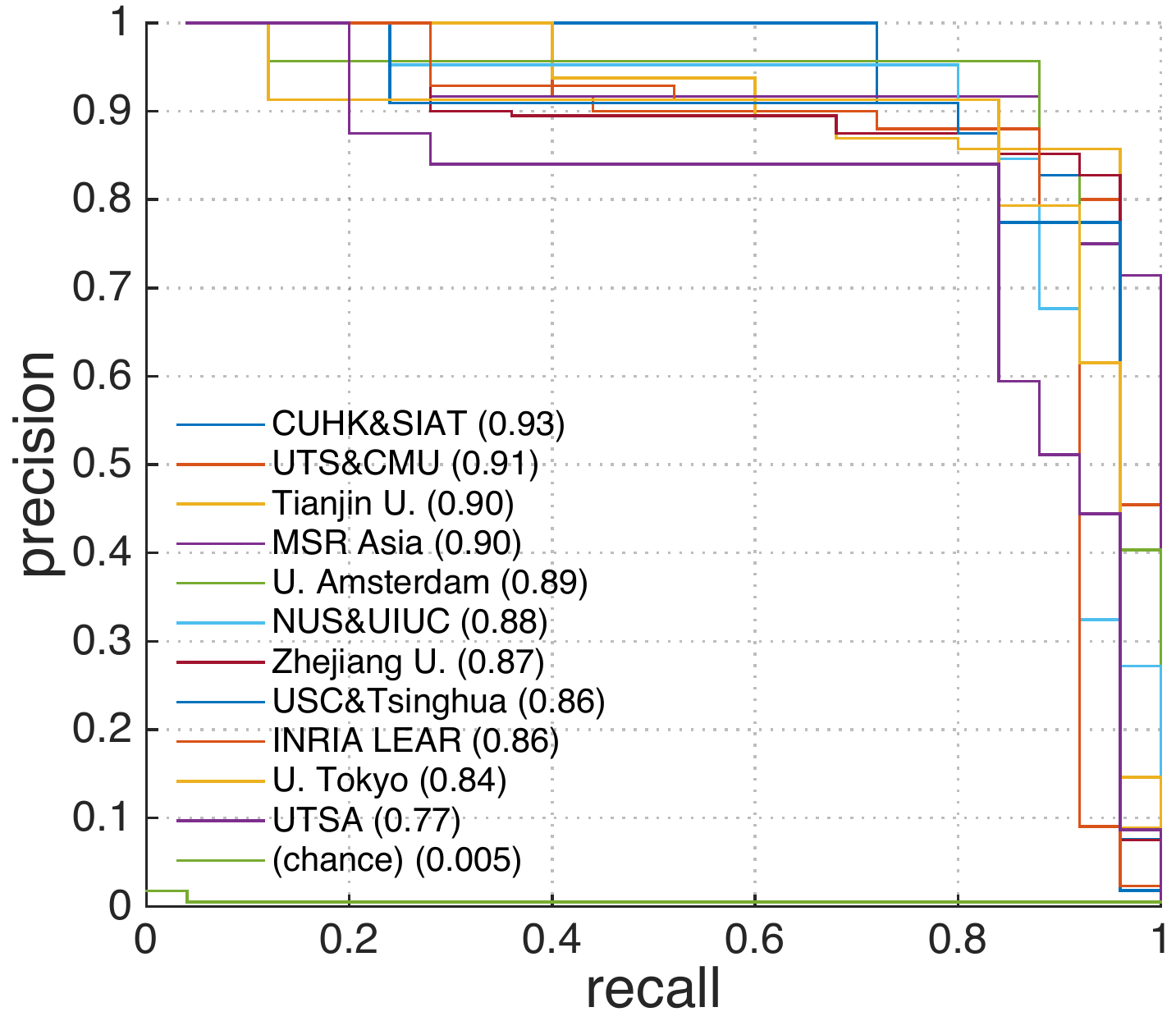}}
   \subfigure[Surfing]{\label{fig:pr_class88}\includegraphics[width=0.44\linewidth]{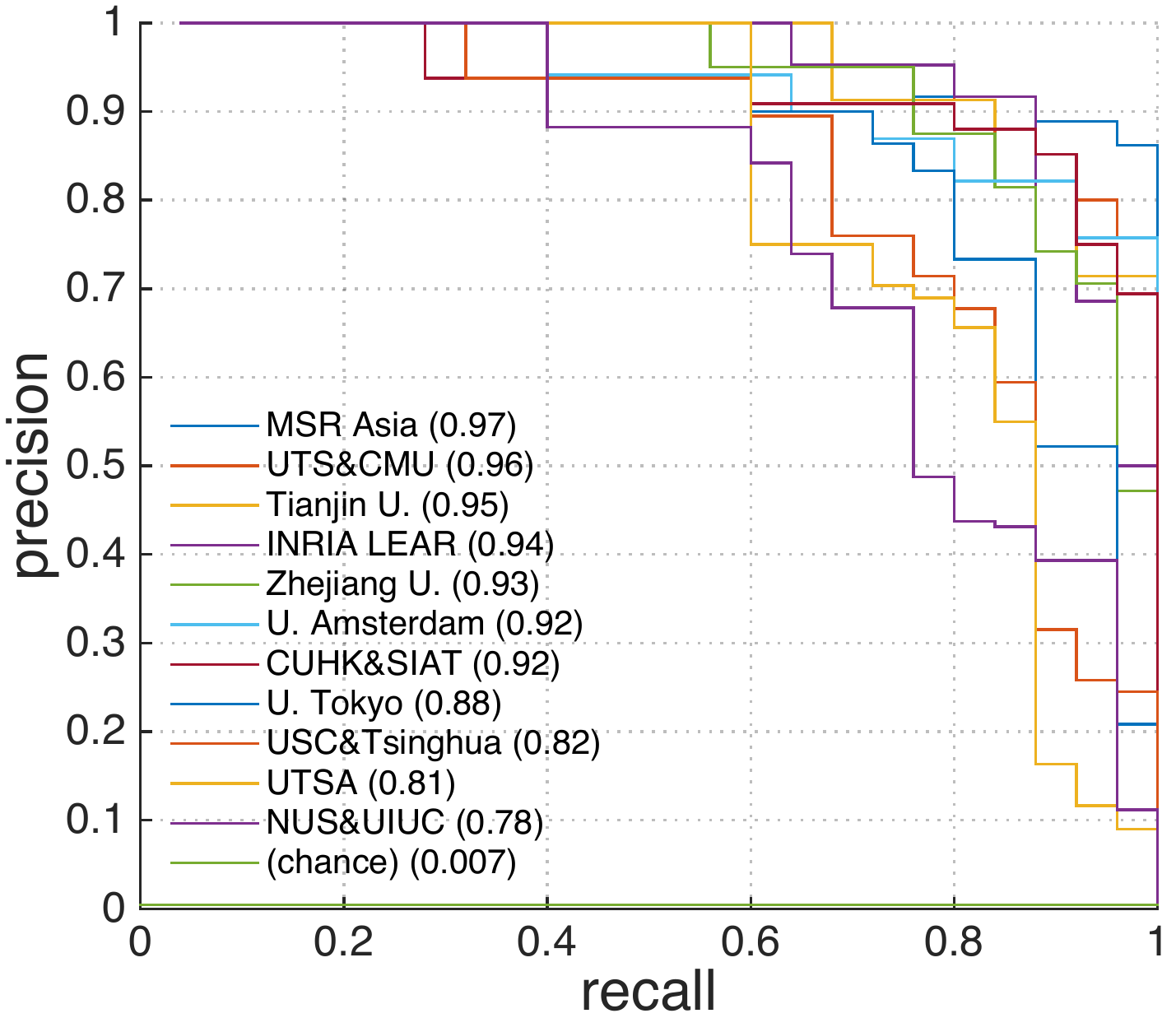}}
   \subfigure[CricketBowling]{\label{fig:pr_class23}\includegraphics[width=0.44\linewidth]{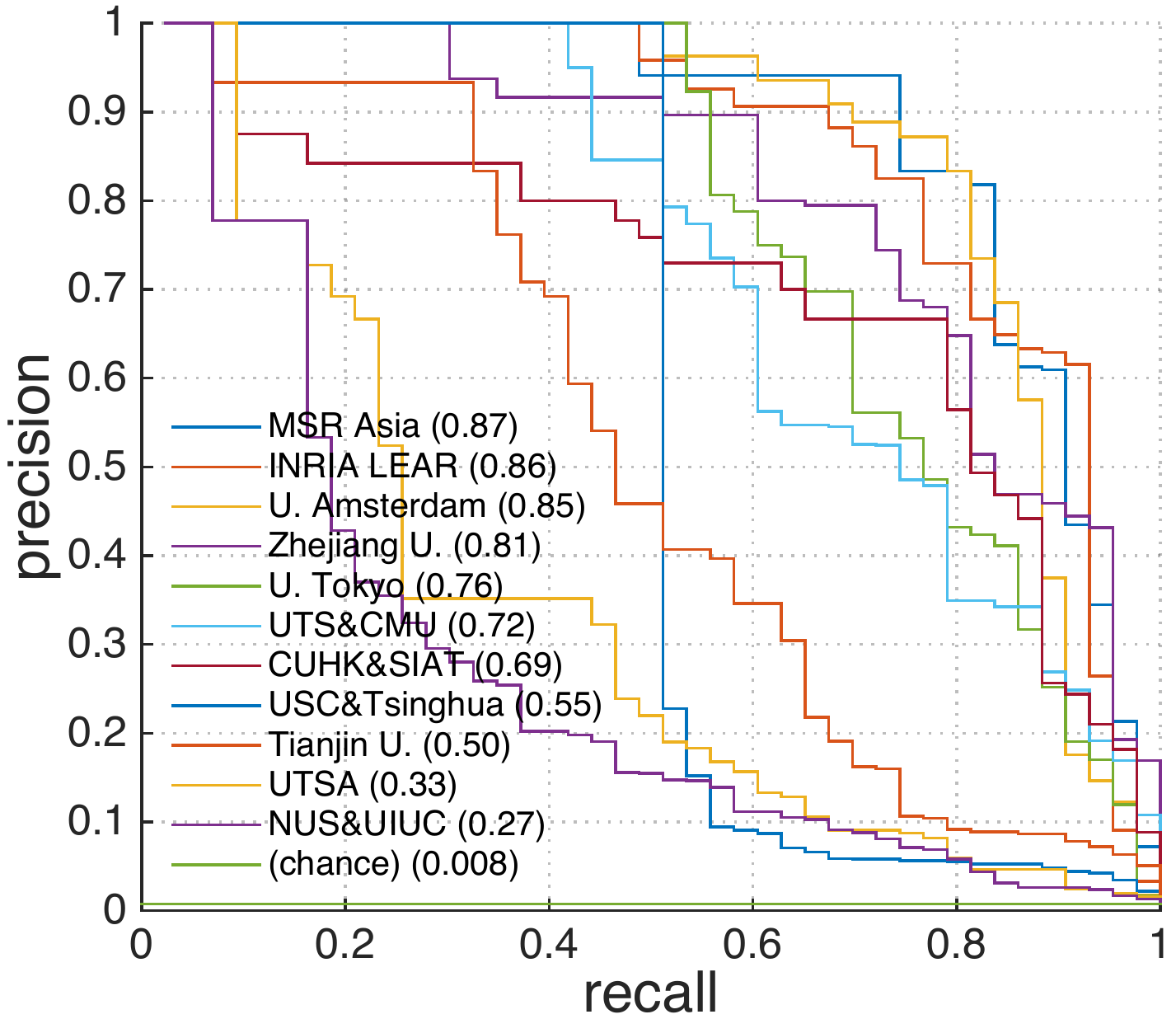}}
   \subfigure[PlayingGuitar]{\label{fig:pr_class63}\includegraphics[width=0.44\linewidth]{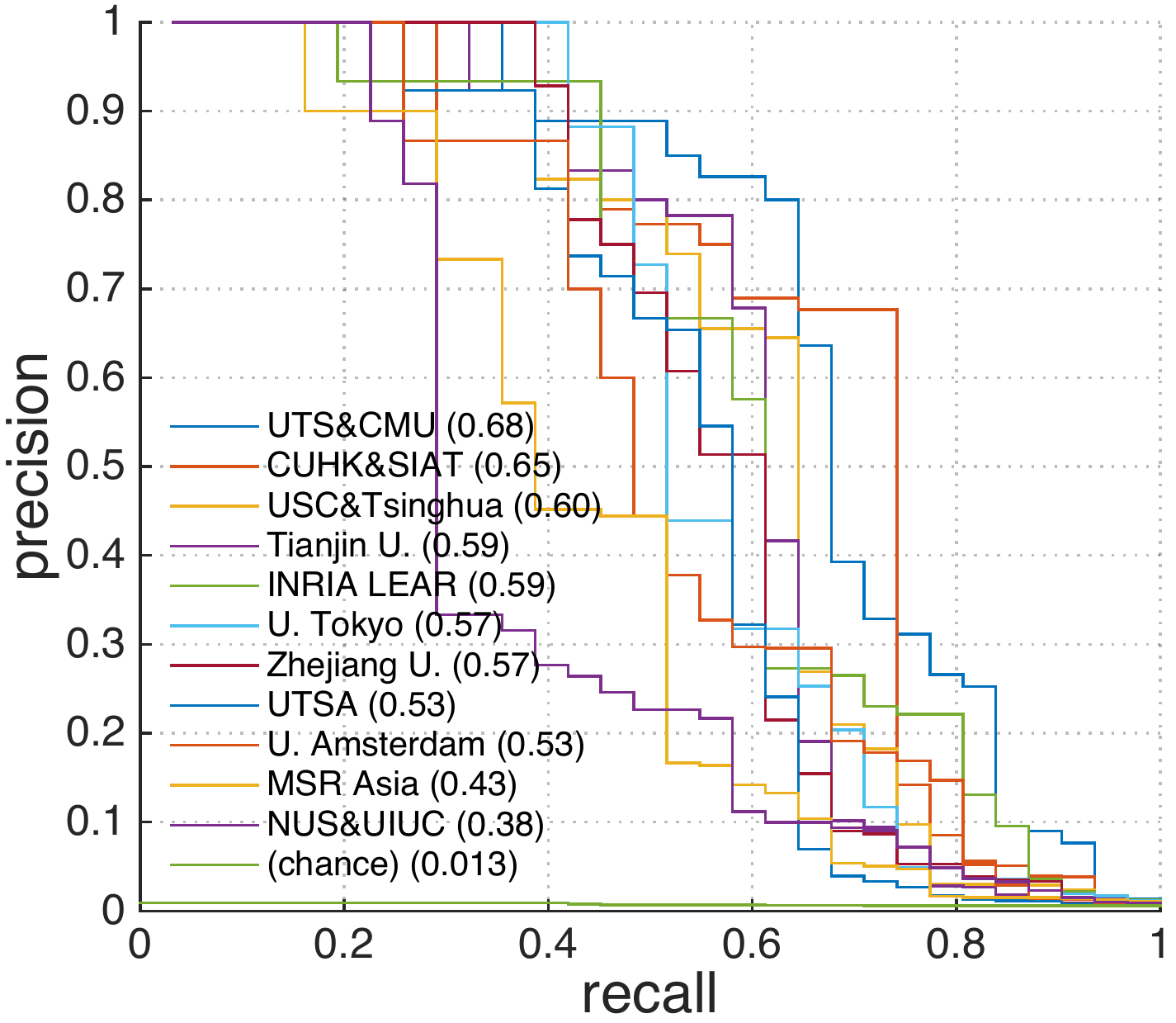}}
   \subfigure[BlowDryHair]{\label{fig:pr_class13}\includegraphics[width=0.44\linewidth]{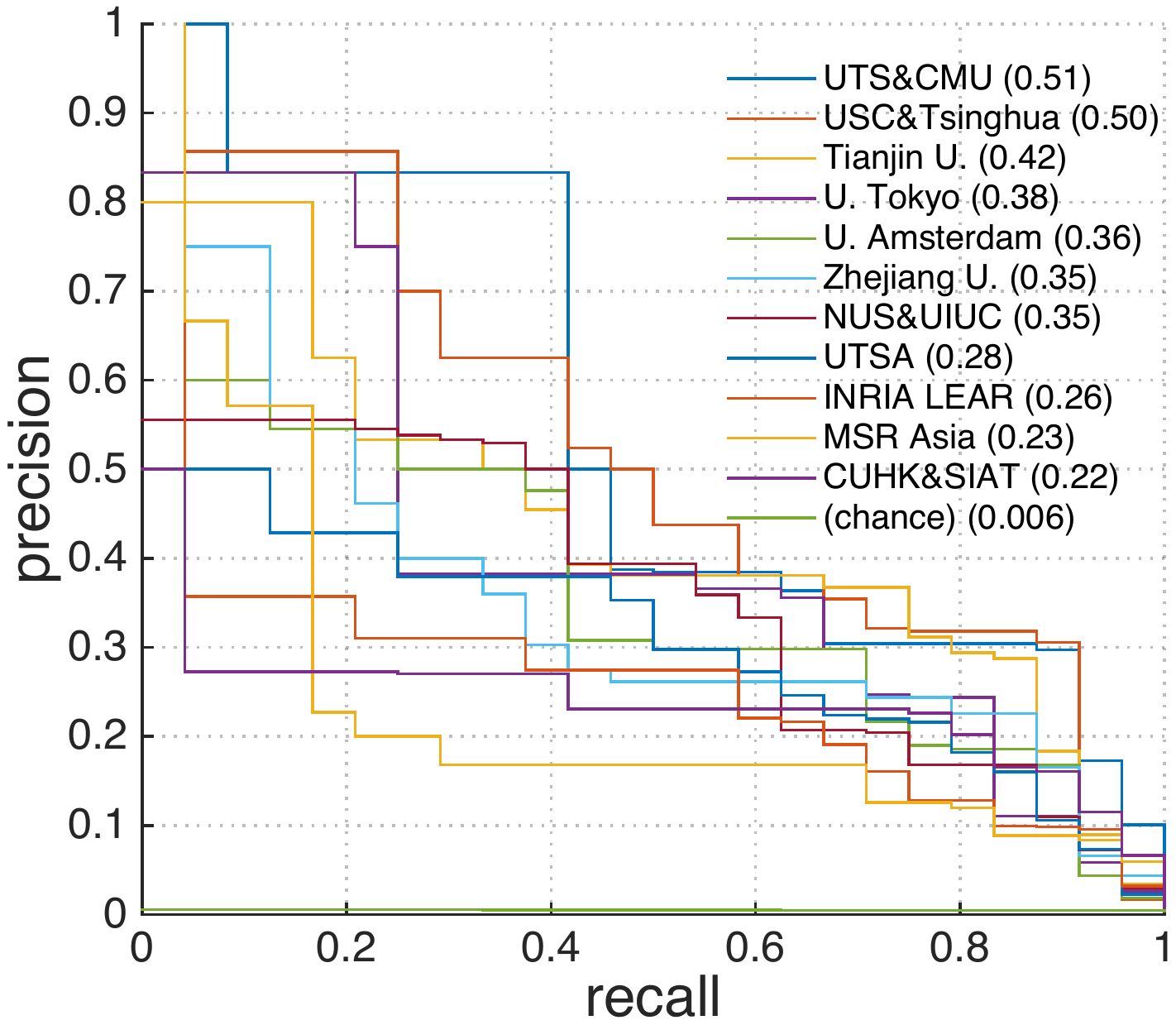}}
   \subfigure[Haircut]{\label{fig:pr_class34}\includegraphics[width=0.44\linewidth]{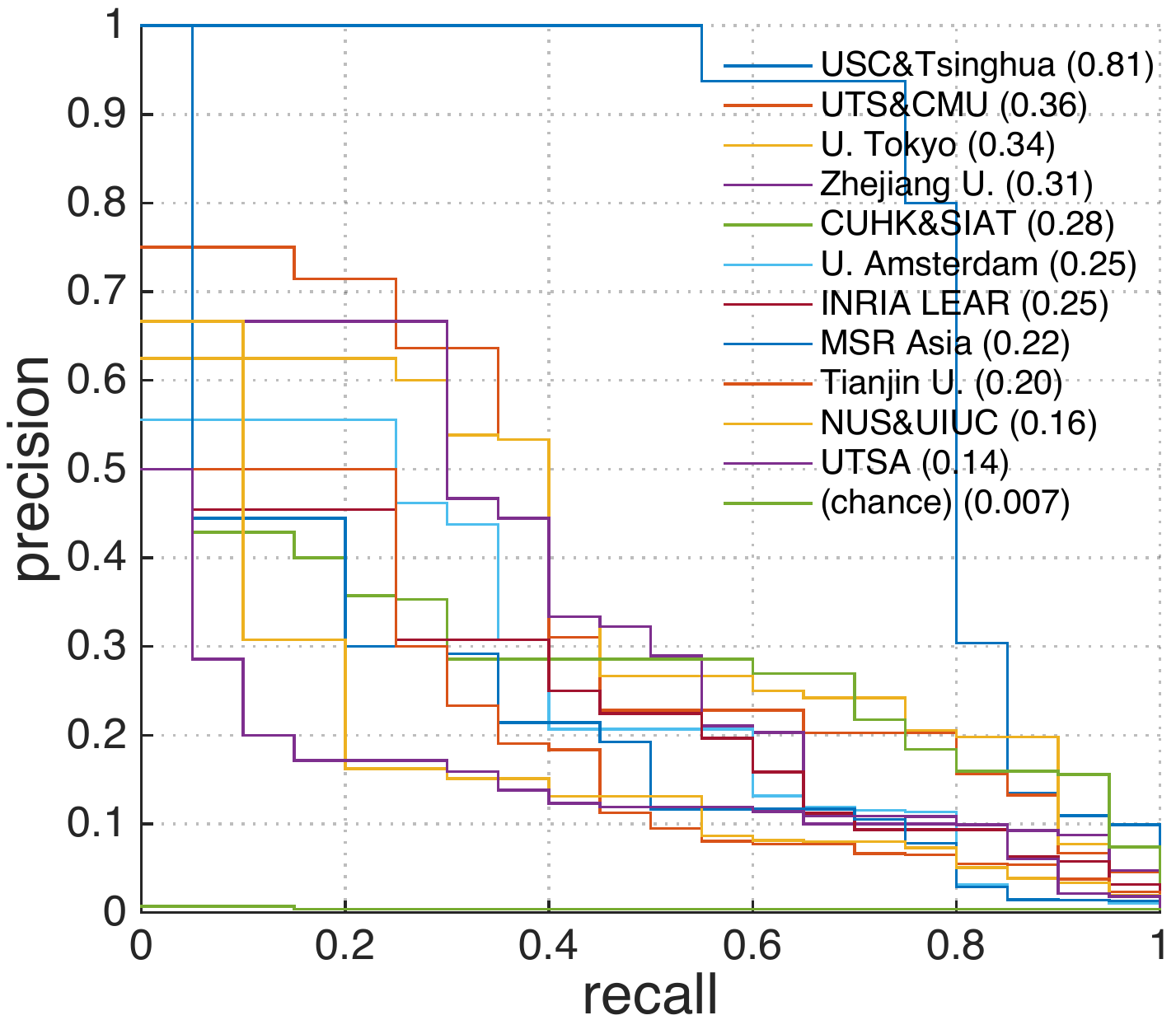}}
    \end{center}
    \caption{Precision-recall curves of a few classes with high (`Bowling', `Surfing'), medium (`CricketBowling', `PlayingGuitar') and low (`BlowDryHair', `Haircut') AP values.}
\label{fig:pr}
\end{figure*}

Figure~\ref{fig:pr} further shows the precision-recall curves. We plot the curves for a few classes with high (`Bowling', `Surfing'), medium (`CricketBowling', `PlayingGuitar') and low (`BlowDryHair', `Haircut') AP numbers. The team names in the legend of each figure are sorted by their AP values. Overall, the classes with higher accuracies tend to contain more unique/representative objects/scenes, while some difficult classes often share similar visual contents that are hard to be separated using state-of-the-art features (e.g., the classes `BlowDryHair' and `Haircut').

\begin{figure*}
\begin{center}
\includegraphics[width=1.0\linewidth]{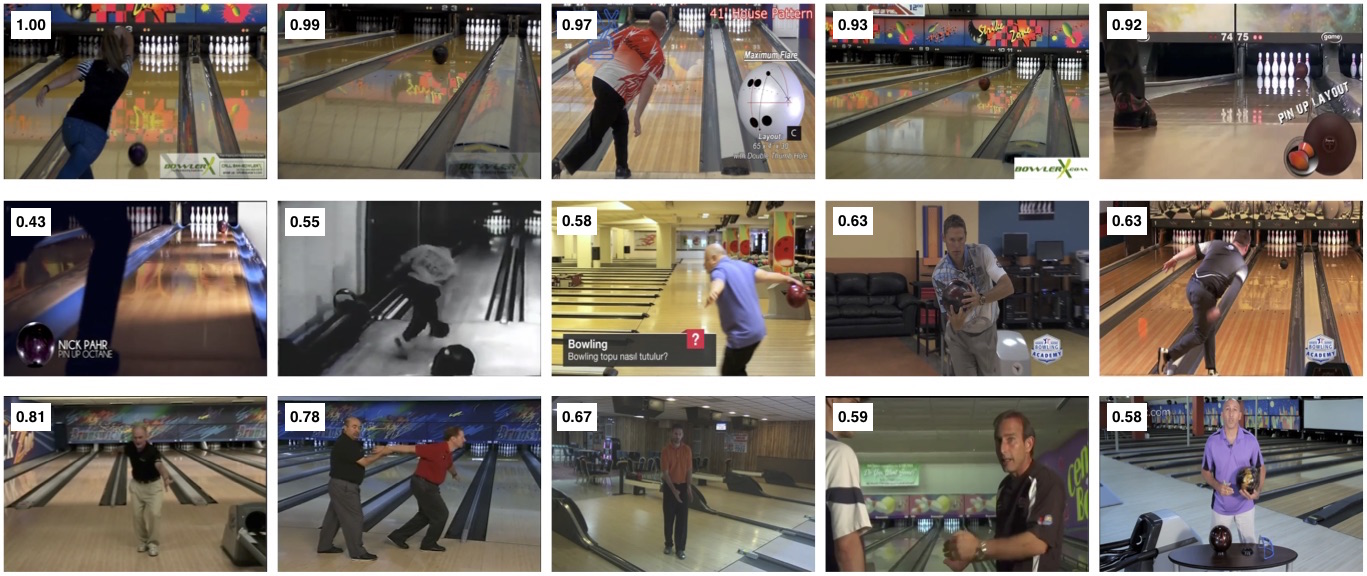}\\
\end{center}
\vspace{-0.8cm}
\caption{Video frames for class `Bowling': First row: top-5 positive videos. Second row: bottom-5 positive videos. Third row: top-5 negative videos. Prediction scores are shown on the frames. }
\label{figExampleBowling}
\end{figure*}

\begin{figure*}
\begin{center}
\includegraphics[width=1.0\linewidth]{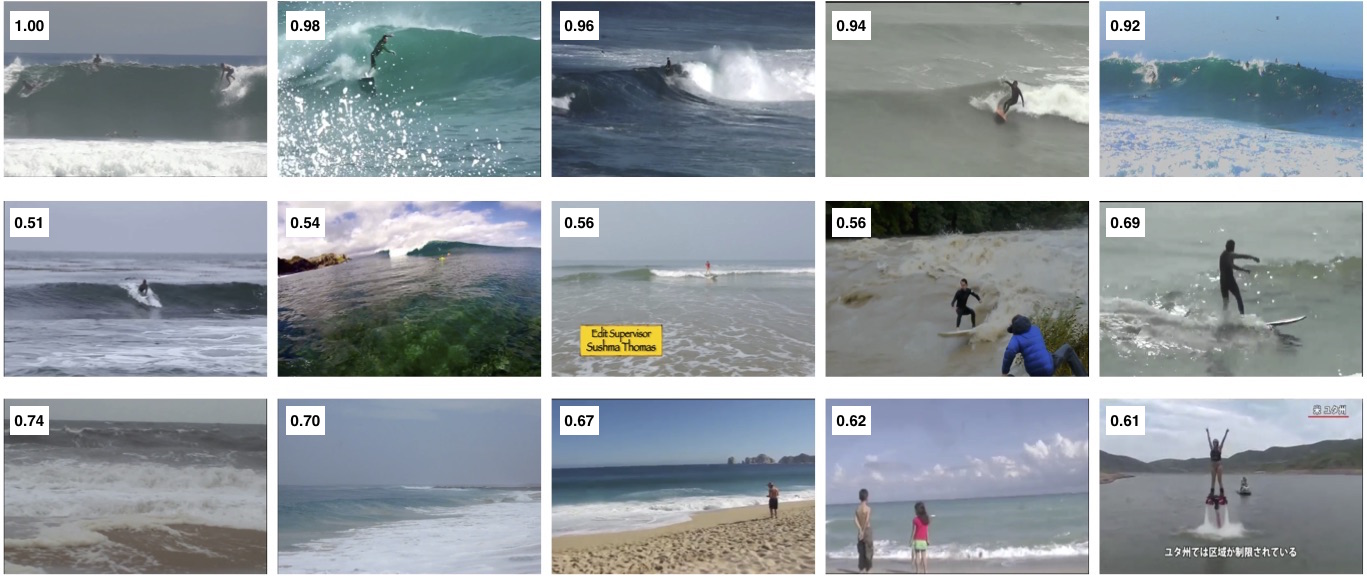}\\
\end{center}
\vspace{-0.8cm}
\caption{Video frames for class `Surfing': First row: top-5 positive videos. Second row: bottom-5 positive videos. Third row: top-5 negative videos. Prediction scores are shown on the frames.}
\label{figExampleSurfing}
\end{figure*}

\begin{figure*}
\begin{center}
\includegraphics[width=1.0\linewidth]{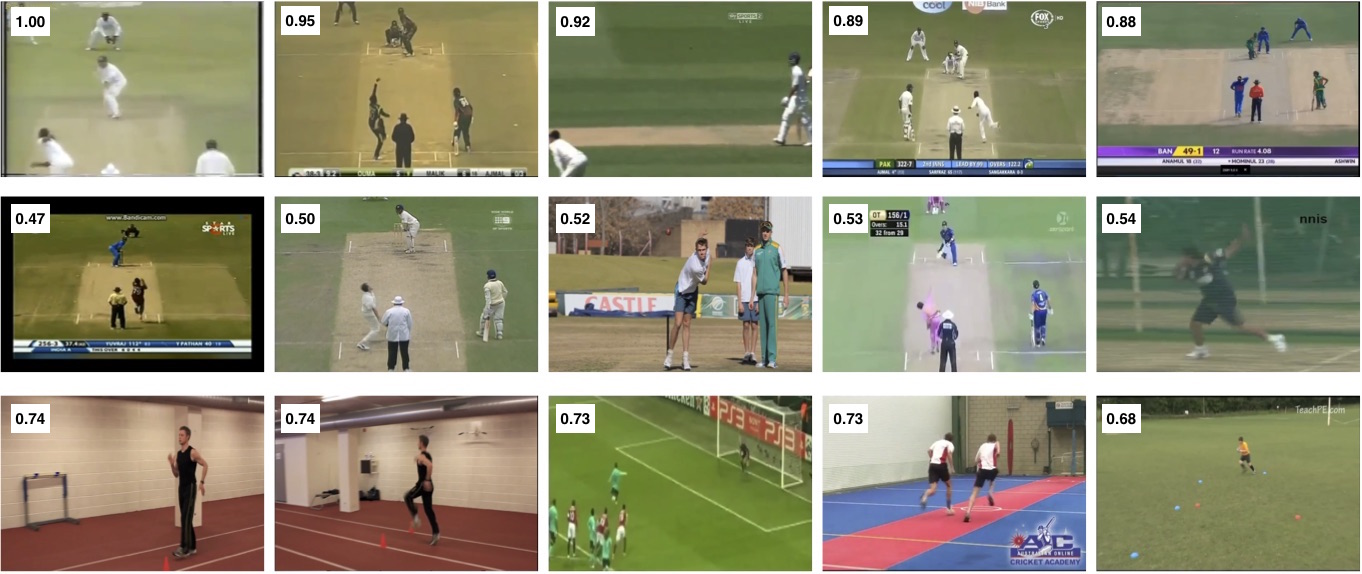}\\
\end{center}
\vspace{-0.8cm}
\caption{Video frames for class `CricketBowling': First row: top-5 positive videos. Second row: bottom-5 positive videos. Third row: top-5 negative videos. Prediction scores are shown on the frames.}
\label{figExampleCb}
\end{figure*}

\begin{figure*}
\begin{center}
\includegraphics[width=1.0\linewidth]{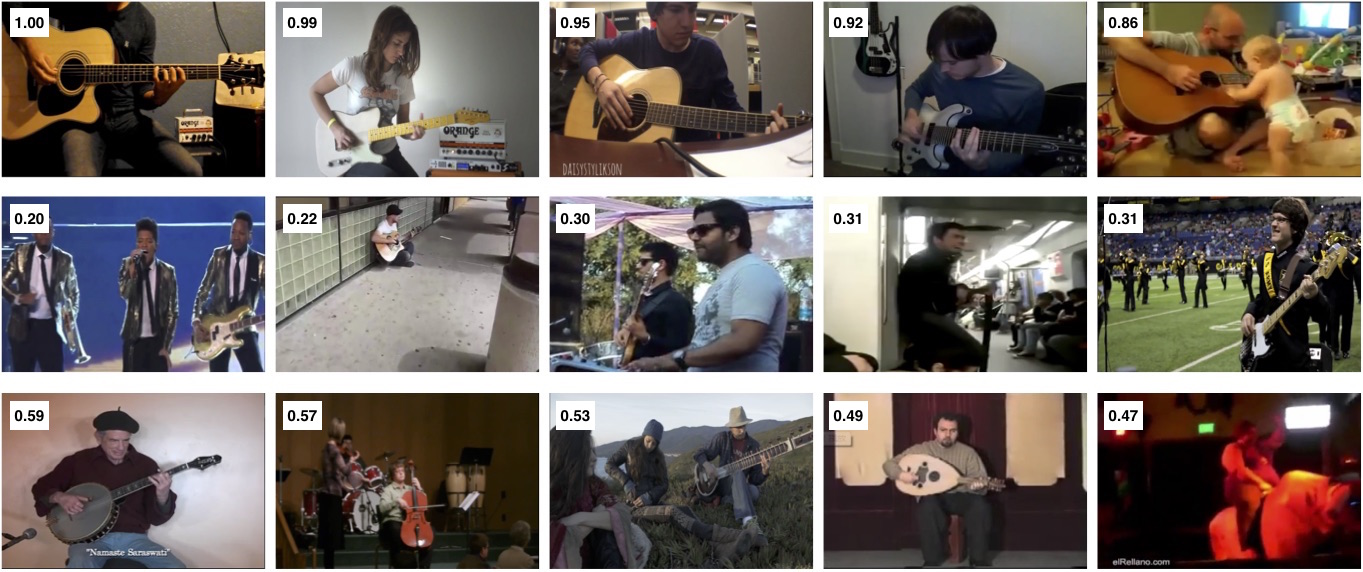}\\
\end{center}
\vspace{-0.8cm}
\caption{Video frames for class `PlayingGuitar': First row: top-5 positive videos. Second row: bottom-5 positive videos. Third row: top-5 negative videos. Prediction scores are shown on the frames.}
\label{figExampleGuitar}
\end{figure*}

\begin{figure*}
\begin{center}
\includegraphics[width=1.0\linewidth]{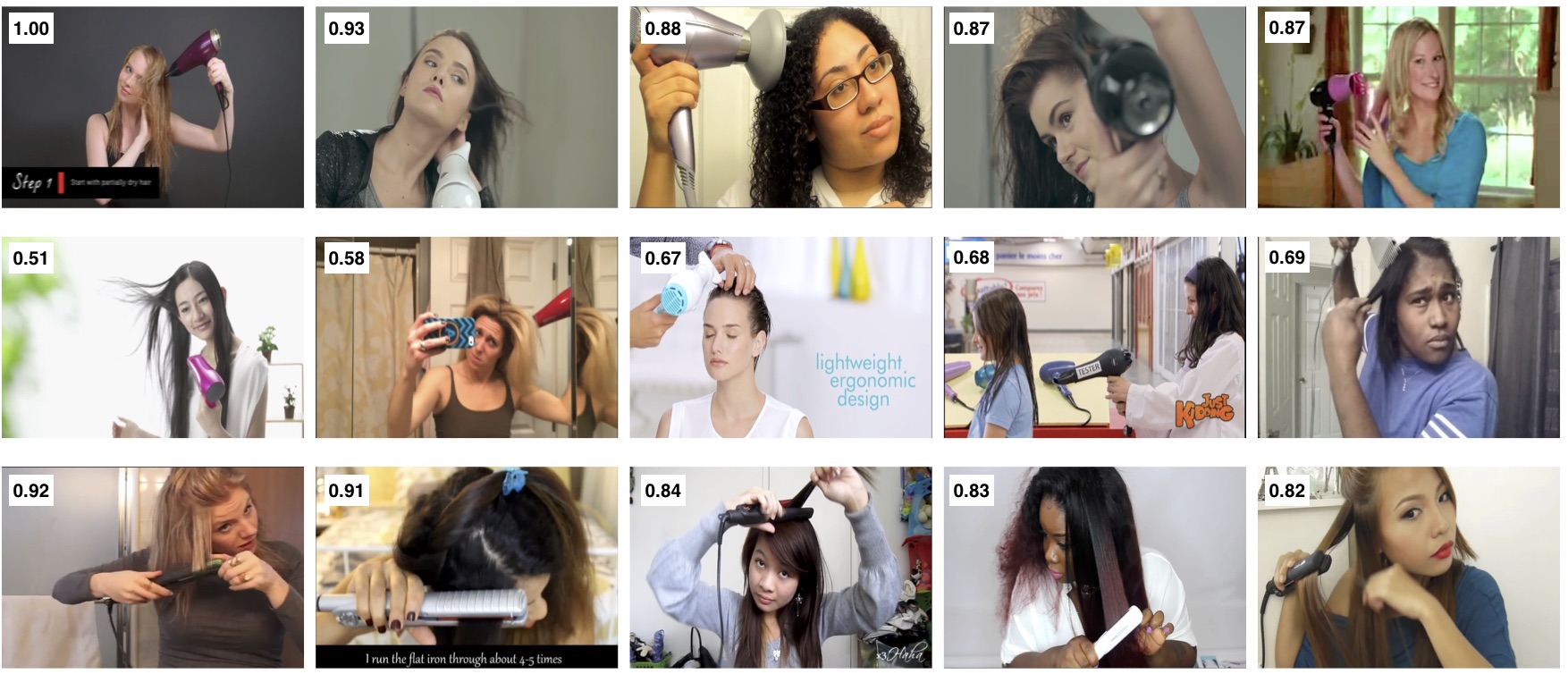}\\
\end{center}
\vspace{-0.8cm}
\caption{Video frames for class `BlowDryHair': First row: top-5 positive videos. Second row: bottom-5 positive videos. Third row: top-5 negative videos. Prediction scores are shown on the frames.}
\label{figExampleBdh}
\end{figure*}

\begin{figure*}
\begin{center}
\includegraphics[width=1.0\linewidth]{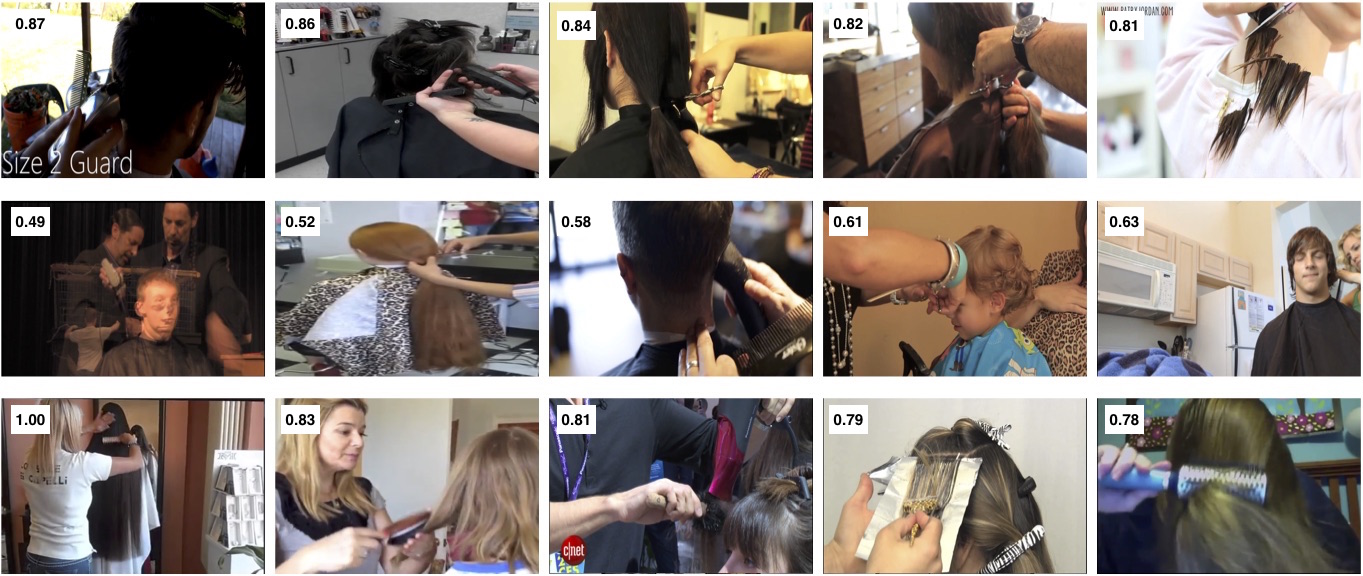}\\
\end{center}
\vspace{-0.8cm}
\caption{Video frames for class `Haircut': First row: top-5 positive videos. Second row: bottom-5 positive videos. Third row: top-5 negative videos. Prediction scores are shown on the frames.}
\label{figExampleHair}
\end{figure*}

We also provide several representative frames from videos in Figures~\ref{figExampleBowling}---\ref{figExampleHair}, respectively for the classes with precision-recall curves shown in Figure~\ref{fig:pr}. The frames are selected based on the best run in THUMOS'15 (from the UTS\&CMU team). For each class, we show the top-5 positive videos found by the best run in the first row, the bottom-5 positive videos in the second row, and the top-5 negative videos (false alarms) in the third row.  As can be seen from the figures, the top ranked negative samples are all visually very similar to the positive ones, which demand more advanced features and classifiers to be correctly separated. We also observe that, for many classes that are easier to be recognized, they contain unique background scene settings. While for the difficult classes (e.g., `BlowDryHair'), the actions may happen under different scene backgrounds. This indicate that current algorithms may significantly rely on background scenes to support action recognition, not just focusing on the actions themselves.

\subsubsection{Impact of Background Videos}
\begin{figure*}[t]
\begin{center}
\includegraphics[width=1.0\linewidth]{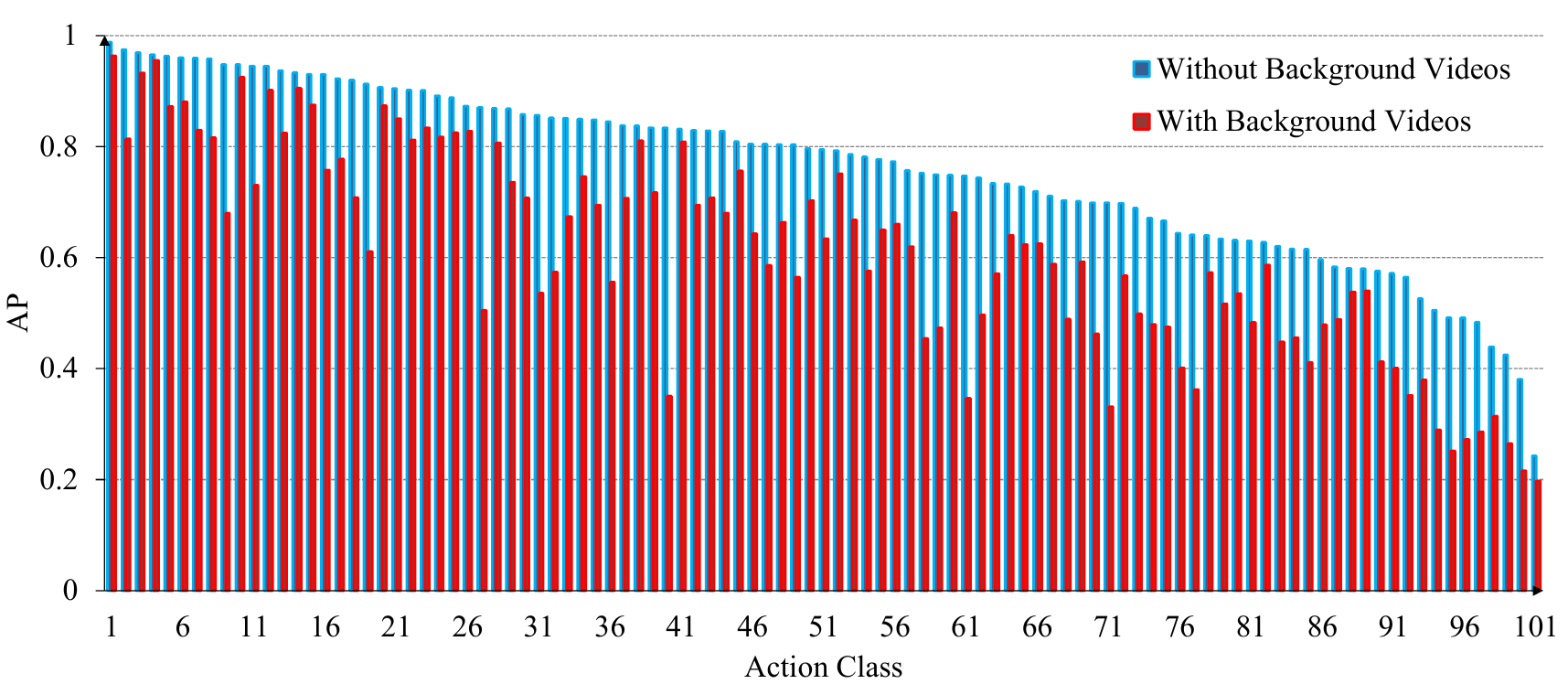}
\end{center}
\caption{Effect of background videos: Blue histogram represents results without the background videos, and red histogram plots results including the background videos. Results are sorted based on the former. This figure is best viewed in color.}
\label{figBgComparison}
\end{figure*}

We also evaluate the impact of background videos in Figure \ref{figBgComparison} which shows AP per-action with and without background videos in the test set. In this figure, the blue histogram represents the results without background videos and the red histogram represents the official results with the background videos. Overall, the mAP after excluding the background videos is 76.3\%, which is 15\% higher than the results with the background videos (61.3\%). This indicates that background videos have critical influence on the performance, which is easy to understand. Some classes like `FrisbeeCatch', `WalkingWithDog' and `BlowDryHair' show significant performance degradation. The main reason is that the background videos contain samples that are visually (but not semantically) similar to these classes. Adding more negative samples during model training might be helpful for these classes. It would be interesting to study this in the future.

\subsection{Temporal Detection}
The results for the temporal detection task for THUMOS'15 are presented in Table~\ref{tb:detection}. In this table, the mAP is computed at overlaps of 10\%, 20\%, 30\%, 40\% and 50\%. Run1 from ADSC, NUS and UIUC has the best results compared to the other four runs, with mAP of $\sim$41\% at an overlap of 10\%. The difference between Run 1 and Runs 2-5 is the use of context features. Run~1 only uses iDT features, while others fuse appearance and scene features from deep networks. This is contradictory to the classification results, where fusion with appearance features in general, and features from deep networks in particular result in significant improvement in performance. However, due to the nature of temporal detection task, the appearance of scene features cause a significant drop in performance. This is because for detection, it is important that the algorithm correctly detects the action, and does not produce false alarms on the rest of the positive videos. The appearance features reduce the discrimination between action segments and background within positive videos, and therefore result in drop in performance. Furthermore, ADSC, NUS and UIUC concluded that it is important to use multiple temporal scales while temporally localizing the actions. Using just a single scale (instead of 9) results in $\sim$30\% drop in performance.

\begin{figure*}[t]
\begin{center}
\includegraphics[width=1.0\linewidth]{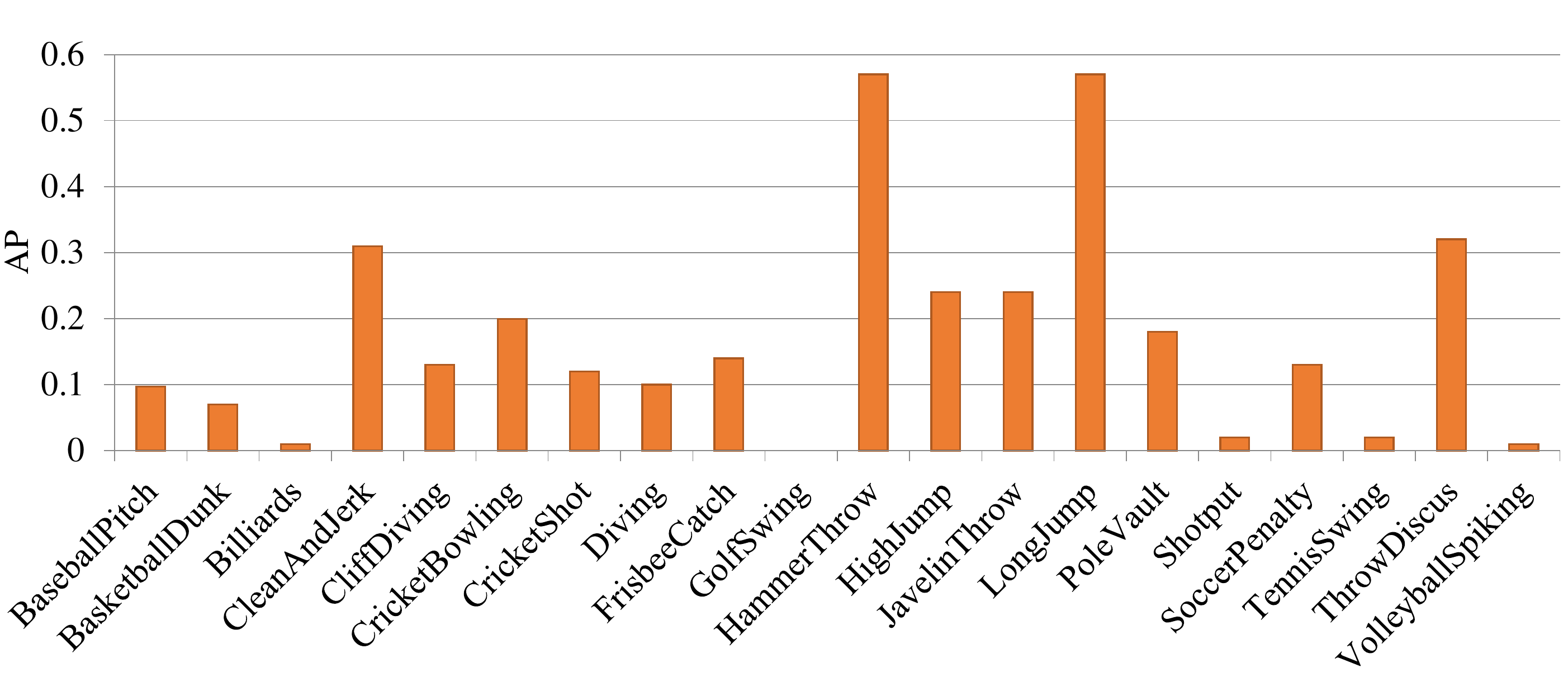}
\end{center}
\caption{Per-Action Average Precision using ADSC, NUS and UIUC - Run1 on the 20 classes used for the temporal detection task.}
\label{figDetectionPerAction}
\end{figure*}

Figure~\ref{figDetectionPerAction} shows the per-action performance on the 20 classes. The action classes with high performance include `HammerThrow', `LongJump', and `ThrowDiscus', whereas the classes with low performance include `Billiards', `ShotPut' and `TennisSwing'. `GolfSwing' and `VolleyballSpiking' have the worse results of all. The results are correlated with the length of the actions, with short and swift actions such as `GolfSwing' being the most difficult to localize.

\renewcommand{\multirowsetup}{\centering}
\begin{table*}[th!]
 \centering
 \small
  \setlength{\tabcolsep}{5pt}
 \begin{tabular}{c||c||c c c c c }
  \hline
  \hline
  \textbf{Rank} & \textbf{Team-Run / Overlap} & \textbf{10\%} & \textbf{20\%} & \textbf{30\%} & \textbf{40\%} & \textbf{50\%} \\
  \hline
  \hline
  1 & ADSC, NUS \& UIUC - Run1  \cite{yuanadsc} & \textbf{0.4086} & \textbf{0.3629} & \textbf{0.3076} & \textbf{0.2351} & \textbf{0.1830}  \\
  1 & ADSC, NUS \& UIUC - Run2  \cite{yuanadsc} & 0.1611 & 0.1349 & 0.1072 & 0.0830 & 0.0562  \\
  1 & ADSC, NUS \& UIUC - Run3  \cite{yuanadsc} & 0.1577 & 0.1346 & 0.1117 & 0.0882 & 0.0652  \\
  1 & ADSC, NUS \& UIUC - Run4  \cite{yuanadsc} & 0.1386 & 0.1154 & 0.0939 & 0.0728 & 0.0510  \\
  1 & ADSC, NUS \& UIUC - Run5  \cite{yuanadsc} & 0.1413 & 0.1180 & 0.0980 & 0.0773 & 0.0552  \\
   \hline
   \hline
\end{tabular}
 \setlength{\tabcolsep}{6pt}
 \caption{Temporal Detection results measured by mAP (\%). Each team can submit up to five runs. The percentages correspond to different values of overlaps. }
\label{tb:detection}
\end{table*}

\section{Action Recognition in Untrimmed Videos}\label{sec:context}

\begin{figure*}[t]
\begin{center}
\includegraphics[width=0.9\linewidth]{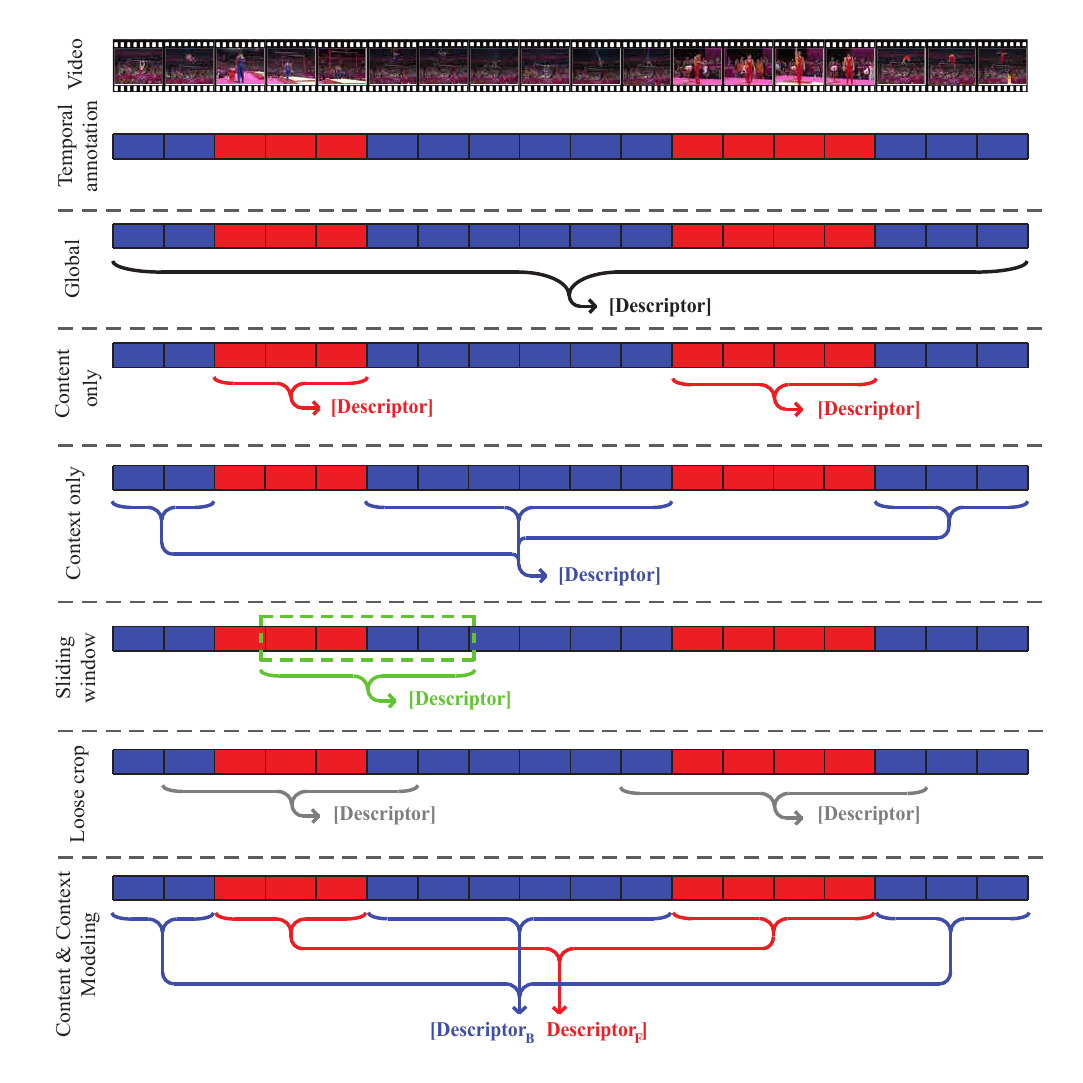}
\end{center}
\caption{Video representations for action recognition in untrimmed videos. Here, red represents a positive action instance in the video whereas blue indicates the background portion.}
\label{figContextDescriptorFormation}
\end{figure*}



The past few decades of research on action recognition has primarily focused on trimmed videos that only contained an action of interest in each video. The lack of a dataset for untrimmed videos and preference of classification over detection task deviated the research on action recognition to focus on pre-segmented trimmed videos. Nevertheless, there have been a few approaches developed for classification~\cite{Bojanowski14,Duchenne09,Karpathy14,Niebles10,Raptis13,Tang12}
and localization~\cite{Hoai11,Pirsiavash14,Ke05,Ke07,Tian13,Yuan09} in untrimmed videos. However, the lack of a large-scale benchmark dataset of untrimmed videos was a pressing need that was first fulfilled in 2014 with the release of THUMOS'14. In this section, we investigate classification performance of state-of-the-art action representations and learning methods in untrimmed setups where target actions occupy a relatively small part of longer videos. In particular, we explore the following questions:

\begin{itemize}
\item What are the important differences between trimmed and untrimmed videos for action recognition?
\item How well methods designed for trimmed videos perform on untrimmed videos?
\item What are the different approaches to represent content and context for action recognition in untrimmed videos?
\end{itemize}

Since we aim to study the role of actions (content) and background (context) in untrimmed videos - which requires temporal annotations - we perform experiments on the 20 action classes with manually annotated action intervals (see Section~\ref{sec:dataset}). Recall that the THUMOS'15 Validation set was formed by merging THUMOS'14 Validation and Test sets, and we collected a new Test set for THUMOS'15. For all the experiments in this section, we used THUMOS'14 Validation Set and/or THUMOS'14 Training Set (UCF101) for training, and the THUMOS'14 Test set for testing.

\subsection{Representations} \label{subsec:representations} To systematically investigate the role of context or background, we construct several representations simulating different amounts of trimming around the action instances (content). These representations are illustrated in Figure~\ref{figContextDescriptorFormation} and are described below: 
\smallskip

\noindent
{\bf R1 - Global:} In the global representation, we extract action descriptor from the full video without using any knowledge about the ground truth action intervals. This is the most straightforward application of standard techniques to untrimmed settings.\smallskip

\noindent
{\bf R2 - Content Only:} Here we assume all action boundaries to be known and extract one descriptor for each action interval. This setup resembles the majority of common action methods and datasets with trimmed action boundaries.\smallskip

\noindent
{\bf R3 - Context Only:} Video intervals outside action boundaries often correlate with temporally close actions and can provide contextual cues for action recognition. For example, tennis swing action co-occurs with running and typically appears on tennis courts. To investigate the effect of contextual cues, we extract descriptors from an entire video excluding action intervals.\smallskip

\noindent
{\bf R4 - Sliding Window:} Here we do not use any knowledge about action boundaries and assume actions occupy compact temporal windows. We model the uncertainty in temporal position of an action and compute descriptors for overlapping windows of length 4 seconds using temporal stride of 2 seconds.\smallskip

\noindent
{\bf R5 - Loose crop:} This setup is derived from the \emph{Content Only} representation by gradually extending the initial action interval into background. We extend initial action boundaries by 1, 3, and 7 seconds before and after the action. Note that the extension of temporal boundaries to the full video is equivalent to the \emph{Global} representation above.\smallskip

\noindent
{\bf R6 - Content \& Context Modeling:} Given a mechanism that can separate content from context, this representation aims to understand if there is any benefit in modeling them separately. Therefore, we combine \emph{Content Only} and \emph{Context Only} representations by concatenating representations computed from action intervals and the temporal background.
\smallskip

\subsection{Features}\label{subsec:features}

Local video features are a standard choice for action representation. We adopt common, standard, and well performing features, in particular Improved Dense Trajectory Features (IDTF)~\cite{Wang13}, to focus on experiments on various representations and methods. Following~\cite{Wang13}, we use HOF and MBH features based on optical flow to capture the motion information in the video. We also use HOG features based on the orientation of spatial image gradients to captures static information in the scene. All descriptors are computed in space-time volumes along 15-frames long point tracks, hence, they capture information in motion-aligned local neighborhood of a video.

To aggregate local features into video descriptors we use Fisher Vector encoding (FV)~\cite{Perronnin10}. FV has been shown to consistently outperform histogram-based bag-of-feature aggregation techniques~\cite{Oneata13}. We use Gaussian Mixture Model with K=256 learned separately for each type of local feature, after reducing the dimensionality of HOG, HOF and MPH using PCA.

Since computing features is the most expensive step to represent video intervals with different temporal locations and temporal extents, we compute FVs for consequent chunks of 10 frames of a video without FV normalization independently for HOG, HOF and MBH. To obtain a FV descriptor for a given video interval, we used the additivity property of Fisher Vectors \cite{oneata2014efficient} by taking weighted sum of FVs corresponding to 10-frames chunks followed by L2 normalization. Thus, this approach allowed us to avoid re-computation of features for generating different representations as required by our setup.

\subsection{Experimental Results}\label{subsec:results}
Next, we report results and analysis of our experiments on action classification and temporal detection in untrimmed videos. We also investigate the role context plays in detecting actions in untrimmed videos. Context refers to the background portion of a positive video which does not contain any instance of the labeled action (R3). We evaluate the different representations in Section~\ref{subsec:representations} to convert the localized (e.g., frame-level) annotations into video-level action labels: \emph{Global, Content Only, Context Only, Sliding window, Loose Crop, and Content \& Context modeling}.

\subsubsection{Action Classification in Untrimmed Videos}

We investigate the first five representations R1--R5 at test time and report action classification results. For training, we assume a fully-supervised setup with known action intervals. We use trimmed videos from the THUMOS'14 Training Set (UCF101) and annotated action instances from the THUMOS'14 Validation Set as positive samples for a particular action class, i.e., one descriptor per positive instance. For negative samples, we generate a single descriptor from each background video in THUMOS'14 Validation Set, and one descriptor per sliding window from the background portion of positive videos (\emph{Context Only}). We learn one-vs-rest classifiers for all action classes, where the negative samples include positive instances from the other classes in addition to background samples. Table~\ref{tab:context_results} summarizes the results of the video-level classification task. For each case, we report the mean average precision, reweighted by the number of instances in each test set. This makes the number of test instances identical for all cases and enables direct comparison between them. We make several observations:

\begin{table*}[t]
\centering
\scalebox{0.9}{
\begin{tabular}{ c||c||c}
\hline
\hline
\textbf{Training Setup}  & \textbf{Testing Representation} &  \textbf{mAP} \\
\hline
\hline
Context Only (R3) & Global (R1) & 0.46 \\
Content Only (R2) & Global (R1) & 0.68 \\
Content Only (R2) & Content Only (R2) & 0.72 \\
\hline
Content Only (R2) & Sliding Window (average pooling) (R4) & 0.77 \\
Content Only (R2) & Sliding Window (max pooling) (R4) & \textbf{0.78} \\
\hline
\hline
\end{tabular}}
\caption{Comparison of the various training and aggregation representations. The mean average precision (mAP) presented is obtained after re-balancing, where we ensure that number of testing instances is identical for all the five cases. This is achieved through repeating each video proportional to the number of action instances contained within that video.}
\label{tab:context_results}
\end{table*}

\begin{figure}[t]
\includegraphics[width=1.0\linewidth]{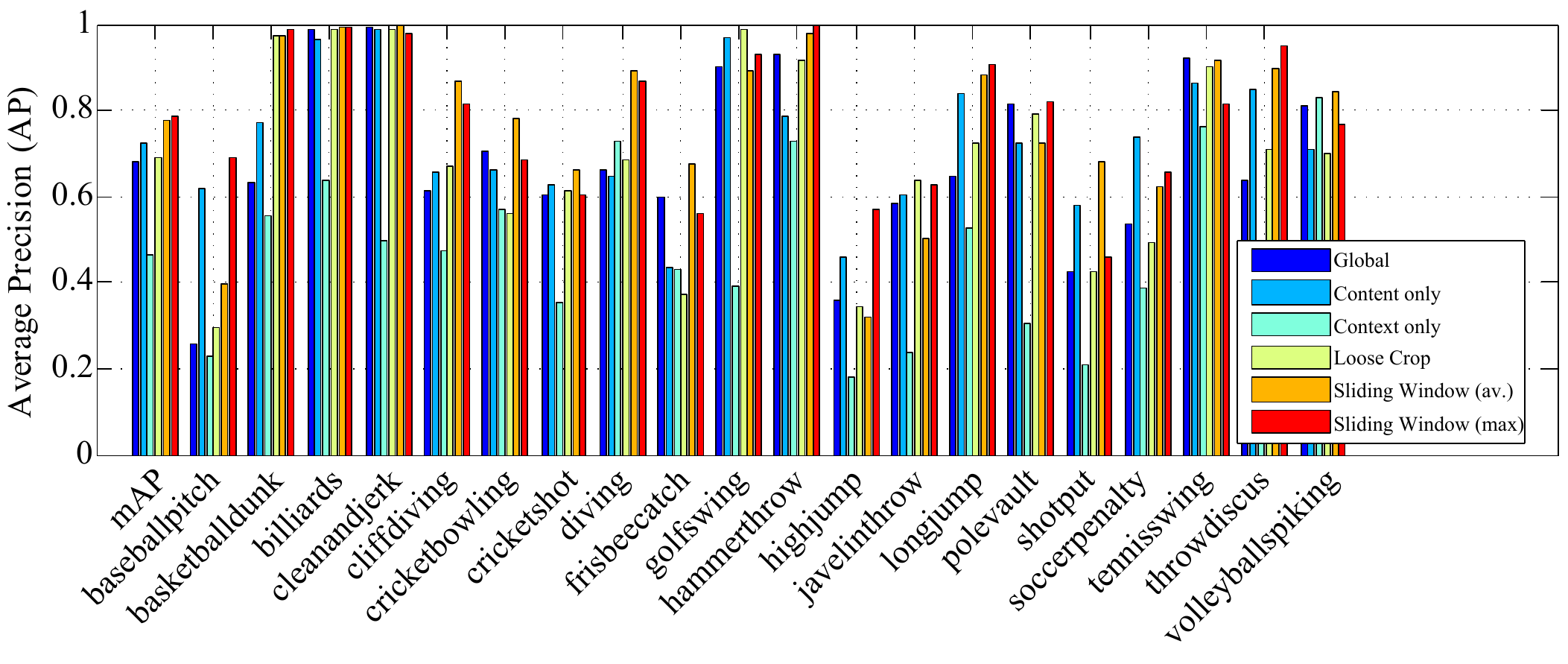}
\caption{Classification Performance: This graph shows the results when training is performed on UCF-101 and trimmed action instances of the THUMOS'14 Validation subset, and testing is performed on THUMOS'14 Test set using representations R1 to R5. \emph{Sliding Window} with both average and max pooling is reported in this graph.}
\label{figContextClassificationResults}
\end{figure}

\begin{itemize}

\item The \emph{Global} case in the second row corresponds to the real-world deployment of a traditional action recognizer, which is trained on trimmed data (\emph{Content Only}) and tested on features aggregated over an entire untrimmed test video. However, comparing this to \emph{Context Only} in the first row is heartening: we confirm that the method is strongly influenced by the frames containing the action of interest (rather than context alone). Removing the action frames drops mAP from 0.68 to 0.46 for IDTF.

\item The \emph{Content Only} in the third row corresponds to the (artificial) scenario, where the action of interest is manually segmented from the untrimmed video, enabling each representation to be aggregated only over relevant frames. 
As expected, mAP improves from 0.68 to 0.72.

\item The \emph{Sliding Window} scenario is a systematic way (though computationally expensive) way to deploy an action recognizer trained on trimmed data on untrimmed videos.  We see that it performs the best and that the choice of pooling strategy (max vs.\ average) has little impact, with max pooling (0.78 mAP) better by only 0.01.

\end{itemize}

\begin{figure*}
\includegraphics[width=1.0\linewidth]{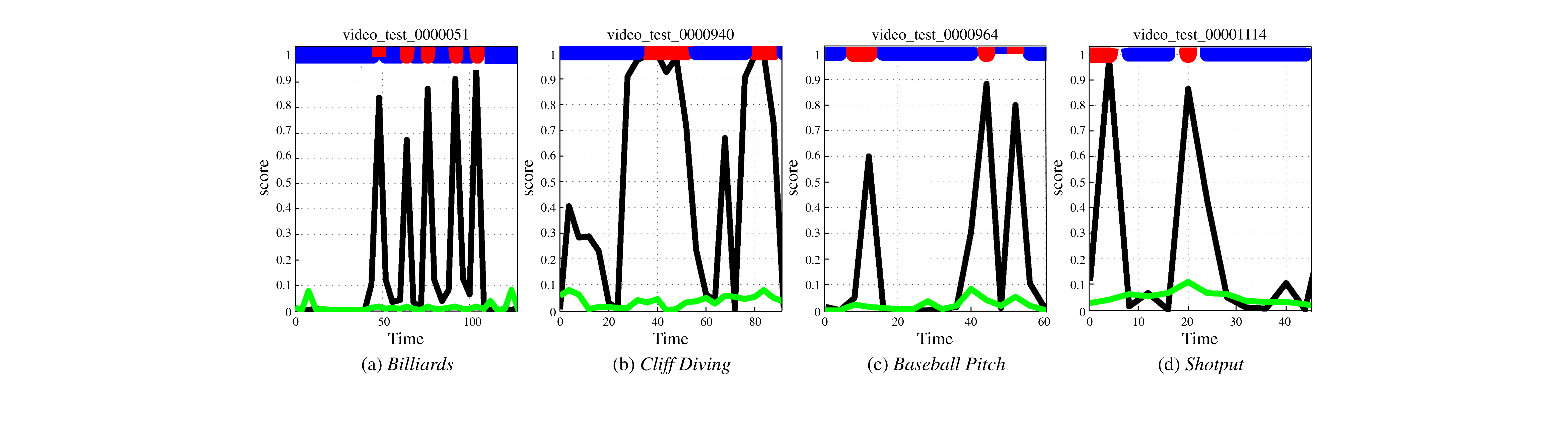}
\caption{Examples of temporal detection scores over time: This figure shows four graphs for different actions. The x-axis is time along the video and y-axis shows the detector scores. The ground truth is shown at the top in red (action) and blue (background). The black curve is detector score for the ground truth action, whereas green curve shows mean of scores from all other detectors. Note that the action of interest rises above the mean response when the action is present, showing why sliding window works well for video-level classification (when pooled) as well as temporal detection.}
\label{figContextQualitative}
\end{figure*}

Figure~\ref{figContextClassificationResults} shows results of these experiments individually for the 20 classes. We also investigate the reason for superior performance of \emph{Sliding Window} approach over other cases. In this regard, Figure~\ref{figContextQualitative} shows examples of temporal detection results for several categories of sample videos. We note that the action of interest (black curve) rises above the average of responses from other actions (green curve) when the action is present. This explains why \emph{Sliding Window} approaches work well for video-level classification with either form of pooling compared to the \emph{Global} representation. The actions are usually much shorter than an entire untrimmed video and the detector gives better performance for those short durations. When aggregated and pooled over multiple smaller windows within the testing video, the overall results improve.

We also performed experiments for different parameters of \textit{Sliding Window} (R4) and \textit{Loose Crop} (R5) with results shown in Table~\ref{tab:context_sliding_window}. For \textit{Loose Crop} experiments in the first five rows, the performance of action classification drops as window length is increased around the action instance. The 120 second loose crop corresponds to the \textit{Global} (R1) case as can be seen with mAP of 0.68 from Table~\ref{tab:context_results}. The results for \textit{Sliding Window} (R4) are shown in the bottom part of Table~\ref{tab:context_sliding_window}. The optimal performance is achieved when the window length is 4 seconds and drops when it is either smaller or larger. This is because the average duration of actions for the 20 classes is around 3.75 seconds, and thus the detector output is optimized around this window length. Nonetheless, the drop in performance is nominal for longer windows and shows \textit{Sliding Window} is not sensitive to window length.

\begin{table*}
  \centering
  \scalebox{0.9}{
  \begin{tabular}{c|c|c|c}
    \hline
    \hline
    \textbf{Testing Representation} & \textbf{Window Length} & \textbf{Pooling} & \textbf{mAP}  \\
    \hline
    \hline
    \multirow{5}{2in}{Loose Crop (R5) \\(1FV per loose GT window)} & 0 sec loose & - & \textbf{0.72} \\ 
     & 1 sec loose & - & 0.71 \\ 
     & 3 sec loose & - & 0.69 \\ 
     & 7 sec loose & - & 0.69 \\ 
     & 120 sec loose & - & 0.68 \\ \hline \hline
    \multirow{8}{2in}{Sliding Window (R4) \\(1FV per sliding window)} & 2 sec long & Max & 0.76 \\ 
     & 2 sec long & Average & 0.77 \\ 
     & 4 sec long & Max  & \textbf{0.78} \\ 
     & 4 sec long & Average & 0.77 \\ 
     & 7 sec long & Max  & 0.77 \\ 
     & 7 sec long & Average  & 0.76 \\ 
     & 10 sec long & Max  & 0.76 \\ 
     & 10 sec long & Average  & 0.76 \\
    \hline
    \hline
  \end{tabular}}
  \caption{Video classification with \emph{Loose Crop} (R5) and \emph{Sliding Window} (R4): For all experiments we train models on UCF101 (1FV per video) + background set (1FV per video) + Validation (1FV per GT window, 1FV for each sliding window on the background part).}
  \label{tab:context_sliding_window}
\end{table*}

\subsubsection{Role of Context for Classification in Untrimmed Videos}

\begin{figure*}
\includegraphics[width=1.0\linewidth]{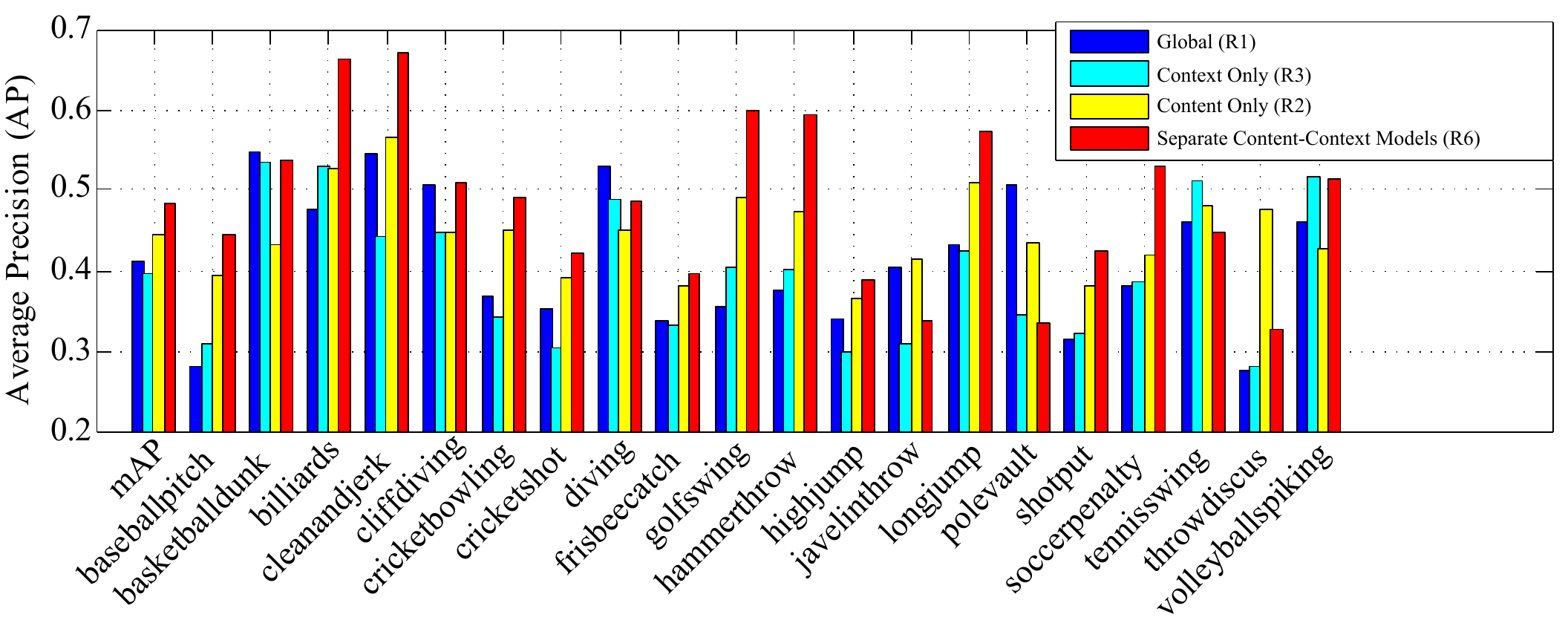}
\caption{This graph shows average precision (AP) for 20 actions using the THUMOS'14 Validation and Test sets for different combinations of content and context for representing the videos.}
\label{figContextEffectResults}
\end{figure*}

Context plays an important role in the ability of the classifiers to make good predictions. However, context alone is not sufficient for obtaining good performance. Removing the action of interest from training decreases performance from 0.68 mAP to 0.46 mAP (Table~\ref{tab:context_results}). The mAP for different runs evaluating the role of context are summarized in Table~\ref{tab:context_localization}, while Fig.~\ref{figContextEffectResults} shows the same for the 20 concepts individually. This particular experiment evaluates on \textit{Content \& Context} (R6) representation and thus the training data requires untrimmed videos containing action instances. Thus, we cannot use UCF101 since its videos are trimmed (no additional context), and background videos from THUMOS'14 Validation set that do not contain content. The training is performed on positive videos from the THUMOS'14 Validation Set, while testing performed on THUMOS'14 Test Set.

In Fig.~\ref{figContextEffectResults}, the blue bars denote the \textit{Global} descriptor for untrimmed videos (R1), light-blue shows \textit{Context Only} (R3), yellow depicts \textit{Content Only} (R2) i.e., trimmed actions, while red marks the results obtained by concatenating descriptors for \textit{Content \& Context} (R6). The graph reveals an important insight that context described separately but used in conjunction with content gives the best performance compared to training using \textit{Content Only} (R2). Therefore, gains in performance can be achieved through separate modeling content and context for action classification. For this run, we used information about action boundaries during testing. In realistic scenario, this is expected to be obtained with methods that can generate generic action proposals.

\begin{table*}
  \centering
  \scalebox{0.9} {
  \begin{tabular}{c|c|c}
    \hline
    \hline
    \textbf{Training Setup}  & \textbf{Testing Representations}  & \textbf{mAP}  \\ \hline \hline
    1FV per GT win, 1FV for each sliding win on BG  & Global (R1) & 0.42 \\
    1FV per GT win, 1FV for each sliding win on BG  & Content only (R2) & 0.45 \\
    1FV per GT win, 1FV for each sliding win on BG  & Context only (R3) & 0.39 \\
    1FV per GT win + 1FV for BG & Content \& Context (R6) & 0.49 \\ \hline \hline
  \end{tabular}
}
  \caption{This table shows the experimental results on different approaches to handling context. The training is performed using positive videos of THUMOS'14 Validation Set and testing is performed on THUMOS'14 Test Set.}
  \label{tab:context_localization}
\end{table*}

\subsubsection{Temporal Detection in Untrimmed Video}

\begin{figure}[tbh]
\includegraphics[width=1.0\linewidth]{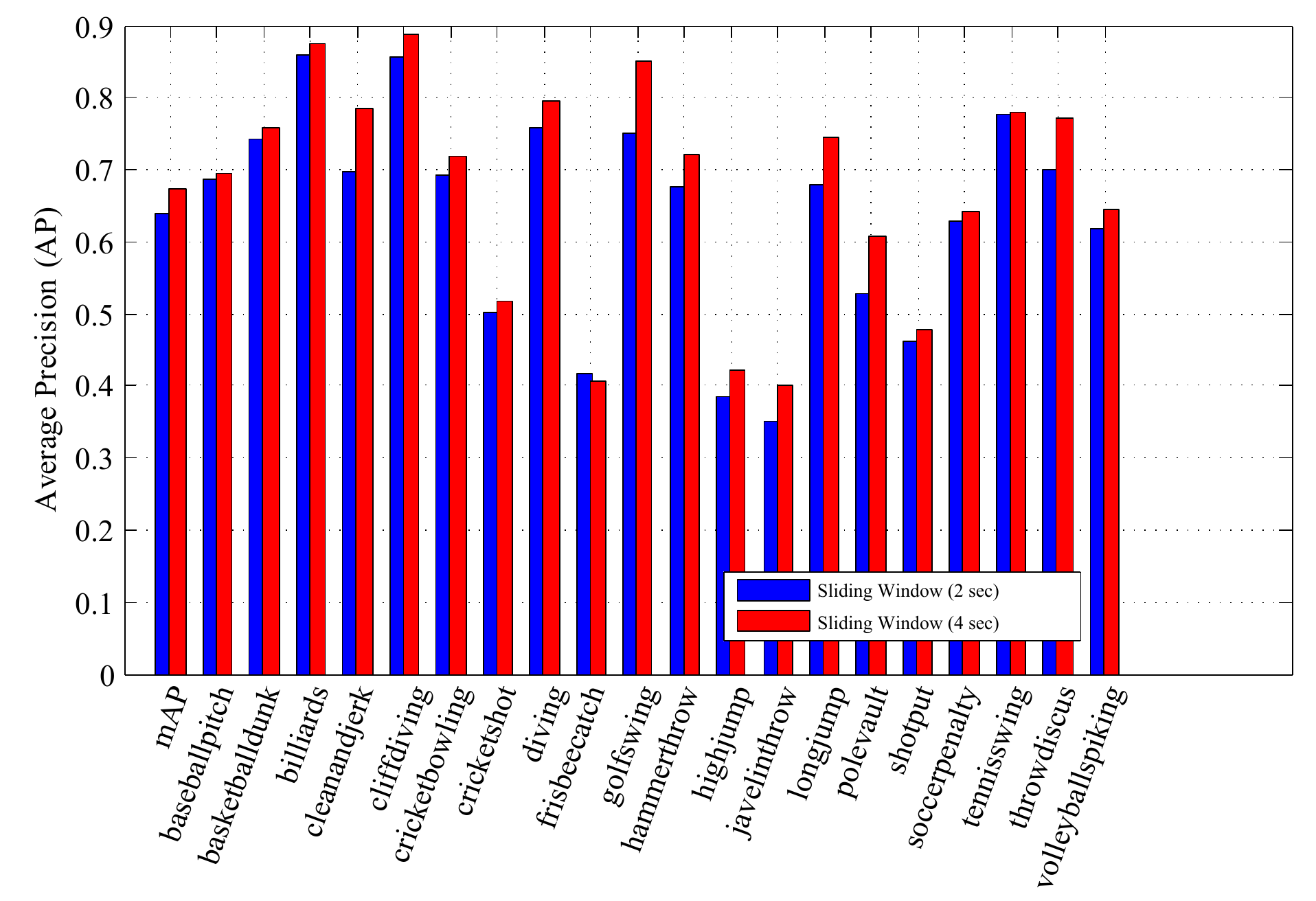}
\caption{This figure shows the temporal detection performance on 20 action classes. The blue and red bars represent sliding windows of 2 and 4 seconds, respectively.}
\label{figContextDetectionResults}
\end{figure}

We also report some results for the task of temporal detection on 20 action classes. In this case, we use the same training setup as for action classification using Training and Validation subsets. At test time we use the classifier in a sliding window manner in combination with temporal non-maximum suppression to select a single action interval for each action hypothesis on the THUMOS'14 Test set.
Fig.~\ref{figContextDetectionResults} reports AP per class using sliding windows. IDTF achieves a mAP of 0.67 on this task. Furthermore, a sliding window for 4 seconds outperforms that of 2 seconds by a margin of 0.03.

\section{Future Directions}\label{sec:future}
There are several thrusts for improving action recognition, we focus on two of them in line with the THUMOS challenge: the dataset and evaluation tasks that quantify performance on different aspects of action recognition. We believe having a denser, more comprehensive, and more generalizable understanding of a video is the way forward. We plan to introduce the \textbf{spatio-temporal localization} task in weakly supervised setting, where training is performed on untrimmed videos without the availability of frame-level annotations or bounding boxes. The test set, however, will contain frame-level and bounding box annotations for the detection and localization tasks, respectively.

THUMOS'15 contains about 13000 trimmed videos for training the 101 action classes, as well as approximately 10500 untrimmed videos in the validation and test sets. The dataset amounts to $\sim$370 gigabytes of data, making it the largest dataset for actions and activities. We propose an \textbf{extension of THUMOS'15} dataset, which despite being the largest video dataset for action recognition, is still deficient both in the number of classes as well as number of instances per class. For that we will define action and activity classes associated with a variety of \textit{verbs}. This will give us the most comprehensive set of classes specifically aimed at capturing human motion. The number of classes will be several times larger than current dataset, with at least 200 instances per action. The space requirement are expected to be on the order of terabytes.

Moreover, our aim is to move from visual (appearance and motion) perception in videos to a deeper semantic understanding by describing different objects, actions and their interaction among themselves and the environment in terms of attributes, semantic relationships and textual descriptions.
Hence, the goal is not only to detect objects and actions, but also explain their complex spatial and temporal interactions. For this, we plan to add a wide variety of videos with primary focus on actions and activities performed by humans, both as individuals or in groups, and then perform \textbf{dense annotations} for objects, actions, scenes, attributes, and the inter-relationships between objects, actions and environment.

For assigning labels to objects, actions and scenes, we propose to use WordNet as it allows modeling of structured knowledge. The WordNet Synsets will relate the different nouns, verbs and adjectives. Here, it will be important to consider the trade-off between \textbf{consistency} and \textbf{diversity}. The consistency requires that we reuse labels that have been used before, so that a particular object or action has the same label across videos. However, this desire is in tension with diversity, as it limits the number of new labels that can be assigned to objects and actions. For instance, the terms `person' and `man' might refer to the same subject. Similarly, the actions `jump' and `plunge' are interchangeable in some contexts. WordNet Synsets are able to relate these words as `person' is a hyperonym of `man', and `jump' and `plunge' are synonyms. Thus, the trade-off could be controlled by preferring specific labels over more general labels and using them consistently, however, other specific labels will be used whenever relevant and available. This will also allow us to transfer many appearance attributes directly from WordNet. The label `grass' will be immediately labeled with green due to the structured knowledge available in WordNet. Indeed, this will require verification from the annotators, but the transfer of attributes and properties will save time and effort while generating richer and dense annotations for a large video dataset.

For cognitive understanding of videos, it is important that training data contains detailed annotations about how the objects, actions and scenes interact with each other. Moreover, qualitative properties of objects and actions, termed \textbf{attributes} also add to the semantic understanding of video data. We will include both \textit{appearance attributes} that capture the visual qualities of objects including color, size, shape, as well as \textit{motion attributes} which are related to the actor, such as the body parts used, their articulation, and type and speed of movement etc. Next, these relationships will be expressed using a structured representation with WordNet. For instance, a man playing violin could be \textit{playing (man, violin)}, and a woman holding eye brush as \textit{holding (woman, eye brush)}. Once these relationships have been constructed for objects, actions, scenes and attributes, they will be merged together to form a graphical representation. The annotators will verify the validity of tree-graphs relating nouns, verbs and adjectives.

The annotations will be supplemented with text, as the ability to produce valid text descriptions of videos is one of the measures of cognitive and high-level understanding. We propose add \textbf{textual descriptions} for all interesting occurrences and events in a video by first annotating with bounding boxes and tubes. Different video regions will have both spatial and temporal overlap with each other, and will have a description of their own. For instance, to be able to detect the action `BasketballDunk', we only need to detect the person performing the action. However, for high-level reasoning such as whether the actor is performing the action independently during practice, or while playing a game with others, it is important that we are able to locate all other objects and detect behaviors of other actors in the video. These dense text captions for each video region will give local summaries and help train better models for cognitive video understanding. The descriptions will be written in third-person present tense, and will be verified for vocabulary and grammatical consistency.

Region-level descriptions in addition to shots selected for summarization through manual annotation will allow evaluation of video-to-text approaches as well. While annotating the videos for descriptions, it is important that the textual summary for regions are not repeated and are diverse enough to delineate the events captured in the video. We propose to do this in an online manner, where new descriptions from an annotator will be nn-gram matched to existing descriptions, and highly matching descriptions will be flagged for an immediate update.

Finally, with the graphical structure representing the objects, actions and attributes in addition to the textual descriptions for regions, it is straightforward to create \textbf{Question and Answer pairs} that go beyond the detection and localization and allow computers to exhibit cognitive understanding. These questions will emphasize the motion of actions, such as:\\

\begin{itemize}[noitemsep,nolistsep]
    \item Which hand did the person use to apply makeup? Which eye?
    \item How long did the person hold the arrow in the bow?
    \item Was the baby crawling on its belly?
    \item What instrument was the person playing?
    \item Where were the people ice dancing?
    \item Who was performing gymnastics?
\end{itemize}




\section*{Conclusion}\label{sec:conclusion}
This paper describes the THUMOS dataset and the challenge is detail. The two tasks include action classification and temporal detection. We gave an overview of the relationship of THUMOS to existing datasets, the procedure used to collect and annotate thousand of videos. Furthermore, we described evaluation metrics used in the challenge and methods and analysis of results for the THUMOS'15 competition. Next, we presented a study on untrimmed videos which were introduced in the 2014 challenge. The results show that sliding window outperforms global description, and separate modeling of content and context is certainly helpful for improving the performance. We also presented several directions to improve the challenge and proposed spatio-temporal localization and weakly supervised action recognition tasks in the future challenges. Finally, by providing a large-scale benchmark dataset of untrimmed videos to the vision community constituting dense annotations of objects, actions and textual descriptions, we hope to foster research in holistic understanding of video data.\\

\noindent\textbf{Acknowledgement:} The authors thank George Toderici (Google Research), Jingen Liu (SRI International) and Massimo Piccardi (Univ.\ of Tech., Sydney) for their contributions to the THUMOS challenge.

\section*{References}
\bibliographystyle{elsarticle-num-names}
\bibliography{thumosCVIU}

\appendix
\section{List of 101 actions} \label{sec:appendixA}
The complete list of actions for UCF 101 and THUMOS is provided below. The actions in \textbf{bold face} are used in the evaluation of the temporal detection task.

\begin{multicols}{3}
\begin{enumerate}[noitemsep,nolistsep]

\item ApplyEyeMakeup
\item ApplyLipstick
\item Archery
\item BabyCrawling
\item BalanceBeam
\item BandMarching
\item \textbf{BaseballPitch}
\item Basketball
\item \textbf{BasketballDunk}
\item BenchPress
\item Biking
\item \textbf{Billiards}
\item BlowDryHair
\item BlowingCandles
\item BodyWeightSquats
\item Bowling
\item BoxingPunchingBag
\item BoxingSpeedBag
\item BreastStroke
\item BrushingTeeth
\item \textbf{CleanAndJerk}
\item \textbf{CliffDiving}
\item \textbf{CricketBowling}
\item \textbf{CricketShot}
\item CuttingInKitchen
\item \textbf{Diving}
\item Drumming
\item Fencing
\item FieldHockeyPenalty
\item FloorGymnastics
\item \textbf{FrisbeeCatch}
\item FrontCrawl
\item \textbf{GolfSwing}
\item Haircut
\item Hammering
\item \textbf{HammerThrow}
\item HandstandPushups
\item HandstandWalking
\item HeadMassage
\item \textbf{HighJump}
\item HorseRace
\item HorseRiding
\item HulaHoop
\item IceDancing
\item \textbf{JavelinThrow}
\item JugglingBalls
\item JumpingJack
\item JumpRope
\item Kayaking
\item Knitting
\item \textbf{LongJump}
\item Lunges
\item MilitaryParade
\item Mixing
\item MoppingFloor
\item Nunchucks
\item ParallelBars
\item PizzaTossing
\item PlayingCello
\item PlayingDaf
\item PlayingDhol
\item PlayingFlute
\item PlayingGuitar
\item PlayingPiano
\item PlayingSitar
\item PlayingTabla
\item PlayingViolin
\item \textbf{PoleVault}
\item PommelHorse
\item PullUps
\item Punch
\item PushUps
\item Rafting
\item RockClimbingIndoor
\item RopeClimbing
\item Rowing
\item SalsaSpin
\item ShavingBeard
\item \textbf{Shotput}
\item SkateBoarding
\item Skiing
\item Skijet
\item SkyDiving
\item SoccerJuggling
\item \textbf{SoccerPenalty}
\item StillRings
\item SumoWrestling
\item Surfing
\item Swing
\item TableTennisShot
\item TaiChi
\item \textbf{TennisSwing}
\item \textbf{ThrowDiscus}
\item TrampolineJumping
\item Typing
\item UnevenBars
\item \textbf{VolleyballSpiking}
\item WalkingWithDog
\item WallPushups
\item WritingOnBoard
\item YoYo
\end{enumerate}
\end{multicols}

\end{document}